%% file: neurips_2026.tex
\PassOptionsToPackage{table,xcdraw}{xcolor}
\documentclass{article}

\PassOptionsToPackage{numbers, compress}{natbib}

 \usepackage[main, final]{neurips_2026}

\usepackage[utf8]{inputenc}
\usepackage[T1]{fontenc}
\usepackage{hyperref}
\usepackage{url}
\usepackage{booktabs}
\usepackage{amsfonts}
\usepackage{nicefrac}
\usepackage{microtype}
\usepackage{xcolor}
\usepackage{multirow}
\usepackage[table]{xcolor}
\usepackage{enumitem}
\usepackage{amsmath}
\usepackage{graphicx}
\usepackage{subcaption}
\usepackage{tcolorbox}
\tcbuselibrary{breakable, skins, listings}
\usepackage{xcolor}
\usepackage{fancyvrb}
\usepackage{algorithm}
\usepackage{algpseudocode}
\usepackage[normalem]{ulem}
\usepackage{wrapfig}

\newcommand{\model}{VTS}

\title{Searching Videos as Trees: Self-Correcting Agents \\ for Grounded Long Video QA}

\author{
Ce Zhang$^{1}$\ \ \
Ziyang Wang$^{1}$\ \ \
Yulu Pan$^{1}$\ \ \
Oluwatumininu Oguntola$^{1}$ \ \ \
\textbf{Pranav Wagh}$^{1}$ \\
\textbf{Qiyu Wu}$^{2}$\ \ \
\textbf{Hiromi Wakaki}$^{2}$\ \ \
\textbf{Mohit Bansal}$^{1}$\ \ \
\textbf{Gedas Bertasius}$^{1}$ \\
$^1$University of North Carolina at Chapel Hill \quad $^2$Sony \\
{\tt\small \{cezhang, ziyangw, yulupan, mohit, gedas\}@cs.unc.edu},\\
{\tt\small \{Qiyu.Wu, Hiromi.Wakaki\}@sony.com}
}

\begin{document}

\maketitle

\input{sections/0_abs}

\input{sections/1_intro}

\input{sections/2_related_works}

\input{sections/3_method}

\input{sections/4_experiments}

\input{sections/5_conclusion}
\input{sections/7_acknowledgements}

\clearpage

\bibliography{main}
\bibliographystyle{plainnat}

\newpage
\appendix
\input{sections/8_supp}

\end{document}

%% file: sections/0_abs.tex
\begin{abstract}
Grounded long-video question answering (Grounded LVQA) requires answering a question about a long video while localizing the short evidence interval that supports the answer. Recent agentic methods frame this task as multi-turn exploration with a single \texttt{crop\_video(start, end)} action, which supports coarse-to-fine narrowing but provides no primitive for fine-to-coarse backtracking. As a result, these agents typically converge prematurely and cannot recover from an early mistake. We propose \textbf{VideoTreeSearch (\model)}, a framework that casts grounded LVQA as iterative self-correcting search over an adaptive temporal tree. \model~constructs a non-uniform tree through adaptive temporal segmentation so that each node corresponds to a semantically coherent segment, and trains an agent to navigate the tree through four discrete operations: \texttt{zoom\_in}, \texttt{zoom\_out}, \texttt{shift}, and \texttt{answer}. These operations expose backtracking and recovery as explicit, learnable primitives rather than implicit behaviors. To train this navigation, we introduce a trajectory synthesis pipeline that produces multi-step paths through the tree, including deliberate detours into incorrect branches followed by recovery. We use these trajectories for supervised fine-tuning, followed by reinforcement learning with grounding and answer-accuracy rewards. On three Grounded LVQA benchmarks (CG-Bench, Haystack-LVBench, Haystack-Ego4D), \model~outperforms the strongest prior multi-turn cropping agents by $+12.5$ mIoU on CG-Bench and $+7.4$ T-F1 on Haystack-Ego4D. The learned policy also transfers to general long-video QA, surpassing all prior multi-turn cropping agents on Video-MME, MLVU, and LVBench by up to $+7.1$ accuracy points. Ablations confirm that self-correcting hierarchical search is the central mechanism behind these gains: removing either adaptive descent or explicit backtracking substantially degrades performance.
Code is available at \url{https://github.com/CeeZh/VTS}.
\end{abstract}

%% file: sections/1_intro.tex
\section{Introduction}
\input{figures/fig_teaser}

Consider an hour-long cooking tutorial paired with the question, ``What did the chef add to the dish right before placing it in the oven?'' Answering it requires the model to first locate the brief moment when the dish is placed in the oven, and then identify what the chef added immediately before. Many real-world long videos share this structure: the evidence needed to answer a question occupies only a small portion of the video, while the surrounding content is largely irrelevant. This task, known as grounded long-video question answering (Grounded LVQA)~\cite{cgbench, lvhaystack}, requires a model to simultaneously localize the supporting evidence interval and produce a final answer. Uniform frame sampling is poorly suited to this setting because sampling too few frames likely skips past the brief evidence interval. In contrast, sampling too many frames exceeds the input budget that current video models can process. Grounded LVQA, therefore, calls for an iterative search, in which the model hypothesizes where the evidence lies and progressively narrows into promising regions.

Recent work formulates this search as agentic interaction with the long video~\cite{yang2025longvt, pan2025timesearchr, zeng2026videoo3, ding2025videozoomer, wang2025avp}. Most of these methods equip a VLM with a single cropping tool, \texttt{crop\_video(start, end)}, and let the agent repeatedly crop the video~\cite{yang2025longvt, pan2025timesearchr, zeng2026videoo3}. While this unified action is flexible in principle, it has two structural shortcomings. First, the action space is asymmetric. Cropping naturally narrows the search but provides no analog for fine-to-coarse backtracking when the agent commits to a wrong video region. Second, continuous timestamp regression provides no hierarchical decomposition of the video. The agent must localize evidence from raw pixels with no prior on where semantically coherent segments lie, leaving the search space flat and unorganized.
As a result, these cropping agents typically converge prematurely and cannot recover from an early mistake.

We propose \textbf{VideoTreeSearch (\model)}, a framework that treats the long video as an adaptive temporal tree and replaces continuous timestamp regression with discrete navigation over its nodes. Our key insight is that organizing the video as a tree of semantically coherent segments, and letting the agent navigate that tree through discrete actions, turns precise localization and backtracking from implicit capabilities into explicitly learnable behaviors. Concretely, \model~first constructs a non-uniform tree whose boundaries are derived from visual and semantic transitions using CLIP~\cite{clip} embeddings, so that each tree node corresponds to a semantically coherent segment. The agent then navigates this tree through four discrete operations: \texttt{zoom\_in} to descend into a child segment, \texttt{zoom\_out} to ascend to the parent, \texttt{shift} to move laterally to a sibling, and \texttt{answer} to commit. This design offers two structural advantages over a flat continuous action space. Because tree nodes are pre-aligned with semantic content boundaries, the model no longer needs to regress precise timestamps from raw pixels, and localization reduces to selecting a node. Exploration and self-correction also become separate primitives. The \texttt{zoom\_in} action supports adaptive coarse-to-fine narrowing, while \texttt{zoom\_out} and \texttt{shift} provide dedicated mechanisms for fine-to-coarse backtracking and self-correction.

Training the agent to use these primitives effectively requires supervision not only for successful navigation but also for recovery from mistakes. Such supervision is not available in existing datasets~\citep{zeng2026videoo3, yang2025longvt, zhang2025thinkingvideos}. We therefore develop a trajectory synthesis pipeline that produces both correct-path trajectories and trajectories that deliberately enter incorrect branches before recovering through ascending and lateral moves. We use these trajectories for supervised fine-tuning, followed by reinforcement learning with outcome-based rewards on grounding and answer accuracy.

We evaluate \model~on three Grounded LVQA benchmarks (CG-Bench~\cite{cgbench}, Haystack-LVBench~\cite{lvhaystack}, Haystack-Ego4D~\cite{lvhaystack}) and three general long-video understanding benchmarks (Video-MME~\cite{videomme}, MLVU~\cite{mlvu}, and LVBench~\cite{lvbench}). \model~substantially outperforms prior multi-turn cropping agents across all three Grounded LVQA benchmarks. Although \model~is trained primarily for evidence localization, its navigation policy also transfers to general long-video QA. \model~surpasses the strongest prior multi-turn cropping agents on Video-MME, MLVU, and LVBench by up to $+7.1$ accuracy points, indicating that tree-grounded search is a useful inductive bias beyond temporal grounding alone. Inspecting the learned policy, we find that \model~averages $4.8$ search turns per video on CG-Bench and invokes a backtracking action in roughly $60\%$ of trajectories across all three benchmarks, indicating that the agent actively relies on its new primitives.

%% file: figures/fig_teaser.tex
\begin{figure*}[t]
    \centering
	\includegraphics[width=0.95\linewidth]{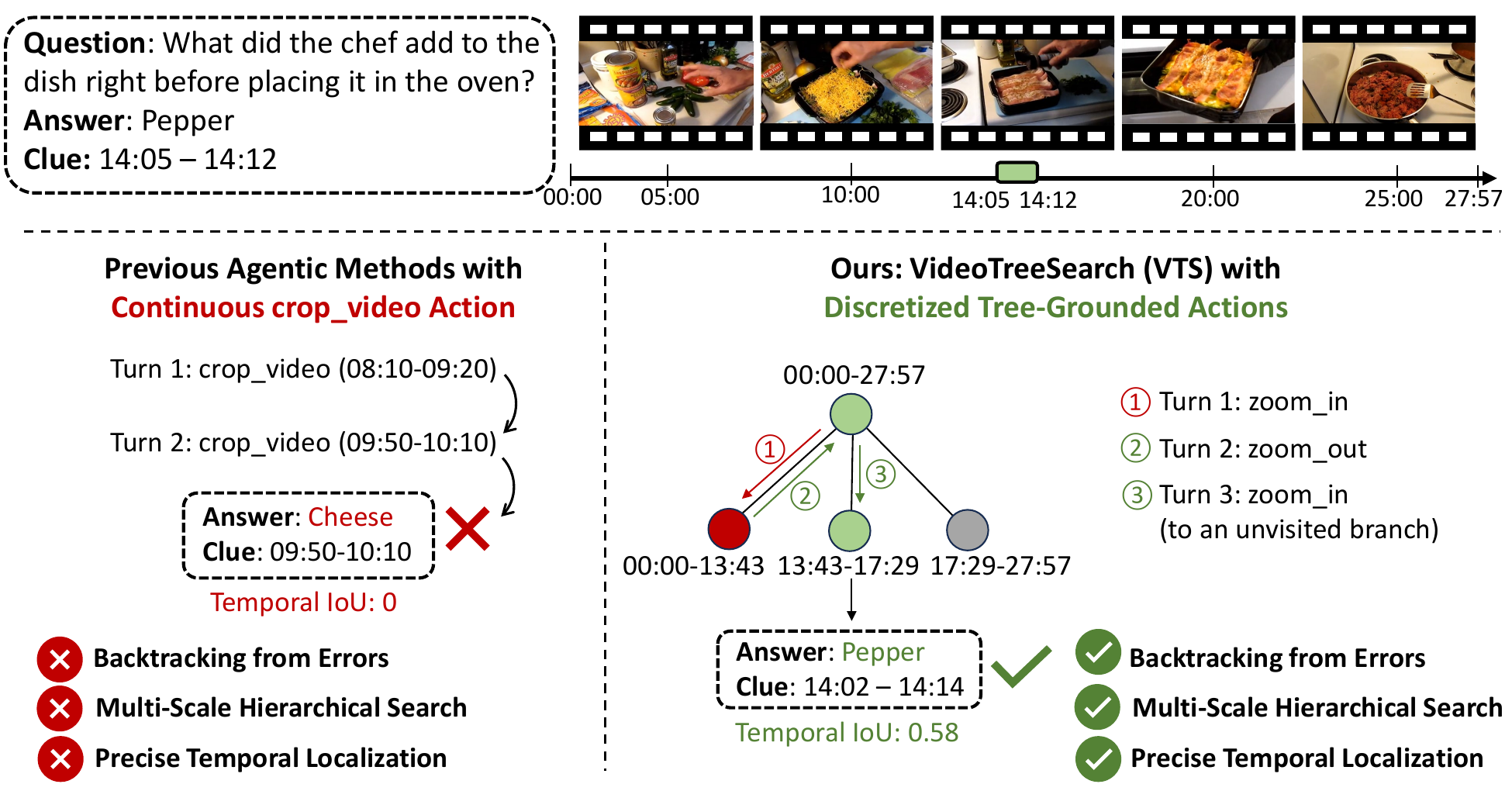}
    \caption{
    \textbf{Comparison between prior multi-turn cropping agents and our VideoTreeSearch (VTS) on grounded LVQA.} Prior multi-turn cropping agents (bottom left) rely on a continuous \texttt{crop\_video} action that conflates exploration with correction. As a result, they often converge prematurely and cannot recover from early mistakes. Our proposed VideoTreeSearch (bottom right) instead organizes the video as a tree of semantically coherent segments and lets the agent navigate it through discrete actions (\texttt{zoom\_in}, \texttt{zoom\_out}, \texttt{shift}). This turns precise localization and backtracking from implicit capabilities into explicitly learnable behaviors.}
    \label{fig:teaser}
    \vspace{-0.1in}
\end{figure*}

%% file: sections/2_related_works.tex
\section{Related Work}

\paragraph{Long video understanding and grounded long-video question answering.}
Long video understanding requires reasoning over videos that span minutes to hours, in which sparse query-relevant cues are buried in mostly redundant content~\cite{videomme, lvbench, mlvu, longvideobench}. Most prior work focuses on the answer side. Video-specific MLLMs process long inputs end-to-end through context extension~\cite{chen2024longvila, zhang2024longva, ren2024timechat, chen2024sharegpt4video}, token compression~\cite{shen2024longvu, wang2025adaretake, zhang2024llamavid, song2024moviechat, ryoo2025xgenmmvid}, keyframe selection~\cite{tang2025adaptive, yao2025generative, park2025toomanyframes, buch2022revisiting}, or state-space backbones~\cite{islam2022long, islam2023scenedetection, islam2025bimba} while captioner-LLM pipelines aggregate clip-level descriptions through a strong language model~\cite{zhang2024llovi, zhang2025silvr, kahatapitiya2025langrepo, ranasinghe2025understanding}. These methods process the video in a single pass and do not localize the supporting evidence. We focus on Grounded LVQA~\cite{cgbench, lvhaystack, yuan2025momentseeker}, a more demanding setting in which the model must produce both an answer and the evidence interval that supports it. The grounding requirement makes predictions verifiable and prevents models from relying on language priors to answer correctly without attending to the relevant evidence. Recent benchmarks such as CG-Bench~\cite{cgbench}, Haystack-LVBench, and Haystack-Ego4D~\cite{lvhaystack} couple grounding with QA, and recent methods including ReVisionLLM~\cite{hannan2025revisionllm}, TimeZero~\cite{wang2025timezero}, VideoChat-R1~\cite{li2025videochatr1}, VideoITG~\cite{wang2025videoitg}, and TimeChat~\cite{ren2024timechat} train VLMs with temporal awareness for fine-grained localization. These methods make a single forward pass or follow a fixed recursive pipeline. We instead cast Grounded LVQA as iterative hierarchical search through a tree of semantically coherent segments, where the agent learns to control its descent across temporal scales and to recover from incorrect choices.

\paragraph{Hierarchical representations of long videos.}
A complementary line of work organizes long videos into hierarchical structures for multi-scale reasoning. VideoTree~\cite{wang2024videotree} clusters dense video features and uses an LLM to organize frames into a query-relevant tree, and VideoMiner~\cite{cao2025videominer} groups frames into a tree and learns keyframe selection via reinforcement learning. Related approaches include Video ReCap~\cite{islam2024videorecap}, which produces captions at multiple temporal scales through recursive aggregation and ReVisionLLM~\cite{hannan2025revisionllm}, which performs coarse-to-fine localization on a fixed schedule. Our framework builds on this hierarchical inductive bias but differs in two key respects. Prior methods consume the hierarchy as a static representation in a single forward pass or fixed recursion schedule, whereas we treat the tree as an interactive environment that an agent navigates over multiple turns, learning when and how to descend. Our agent can also re-enter previously visited nodes through ascending and lateral moves, providing explicit self-correction that fixed recursive pipelines lack.

\paragraph{Temporal video segmentation.}
Temporal video segmentation aims to partition a video into coherent temporal units. One line of work focuses on shot boundary detection, which detects editing transitions such as hard cuts and gradual transitions between consecutive shots~\cite{soucek2020transnetv2, zhu2023autoshot}. Recently, OmniShotCut~\cite{wang2026omnishotcut} formulates shot boundary detection as relational prediction with a shot-query Transformer.
Another line of work groups shots into semantically coherent scenes~\cite{chen2021shotcol, wu2022scrl, islam2023scenedetection}, or detects generic event boundaries that humans naturally perceive~\cite{shou2021gebd, zheng2024fgdn, hwang2025diffgebd}.
However, these methods produce a single flat partition of the video at a fixed granularity and are optimized for boundary accuracy.
Unlike these methods, our adaptive temporal segmentation recursively partitions any segment selected by the agent to build a multi-scale tree for search, while the final \texttt{answer} action predicts the precise evidence interval.

%% file: sections/3_method.tex
\section{Method}
\label{sec:method}

\input{figures/fig_method}
We present \model, a framework that frames Grounded LVQA as iterative hierarchical search over an adaptive temporal tree. An overview of the method is shown in Figure~\ref{fig:method}. Given a long video $V$ and a question $Q$, \model~must produce both a final answer $\hat{y}$ and the evidence interval $[t_s, t_e] \subseteq [0, |V|]$ that supports it. The evidence interval is typically only seconds to tens of seconds wide, while the surrounding content is irrelevant. \model~constructs a non-uniform temporal tree whose nodes correspond to semantically coherent video segments (\S\ref{sec:tree_construction}). An agent then navigates this tree through four discrete actions that expose hierarchical descent and self-correction as explicit moves (\S\ref{sec:action_space}). A trajectory synthesis pipeline produces navigation paths with deliberate wrong-branch detours followed by recovery (\S\ref{sec:trajectory_synthesis}), and these trajectories are used for supervised fine-tuning followed by reinforcement learning (\S\ref{sec:training}).

\subsection{Adaptive Temporal Tree Construction}
\label{sec:tree_construction}

The temporal tree exposes the multi-scale structure of the video to the agent. The root corresponds to the entire video, intermediate nodes correspond to coarse-grained segments, and leaves correspond to fine-grained candidates for the evidence interval. Children of a node form a non-overlapping partition of their parent's interval, so descending the tree progressively narrows the agent's hypothesis. Rather than imposing a fixed branching factor or uniform splits, we construct each node's children adaptively, placing boundaries where the visual content changes most.
This design ensures that descent through the tree corresponds to refinement along natural content boundaries rather than arbitrary partitions.

We refer to this procedure as adaptive temporal segmentation. Concretely, given a segment $S = [t_{\text{start}}, t_{\text{end}}]$, we sample up to $64$ frames uniformly from $S$ at $1$ fps and compute a CLIP~\cite{clip} embedding $e_i$ for each sampled frame. We compute the cosine distance $\delta_i = 1 - \cos(e_i, e_{i+1})$ between consecutive embeddings as a measure of visual and semantic change, and place boundaries at frames where this distance exceeds an adaptive threshold $\tau = \text{mean}(\delta) + k \cdot \text{std}(\delta)$, using the later frame of each transition as the boundary. We constrain the number of children to between $3$ and $8$, retaining the top boundaries when the threshold yields too many splits and relaxing the threshold when it yields too few. We stop splitting when the current node is shorter than $64$ seconds, so nodes below this duration become leaves of the tree. The tree is constructed lazily during inference: only the root and the children of nodes that the agent actually visits are materialized. This keeps the navigation space small and ensures that segmentation reflects the regions the agent has chosen to inspect.

\subsection{Tree-Grounded Action Space}
\label{sec:action_space}

At each turn, the agent occupies a node of the tree and selects one of four discrete actions:
\begin{itemize}[leftmargin=*, itemsep=2pt, topsep=2pt, parsep=0pt]
    \item \texttt{zoom\_in($c$)}: descend into the $c$-th child of the current node.
    \item \texttt{zoom\_out()}: ascend to the parent of the current node.
    \item \texttt{shift($s$)}: move laterally to the $s$-th sibling under the same parent.
    \item \texttt{answer($\hat{y}, [t_s, t_e]$)}: terminate with a final answer and evidence interval.
\end{itemize}
A single \texttt{zoom\_in} action descends one level rather than committing to a specific timestamp range, so reaching a fine-grained evidence interval naturally takes multiple turns of progressive narrowing. Exploration and self-correction are decoupled: \texttt{zoom\_in} advances the search downward, while \texttt{zoom\_out} and \texttt{shift} are dedicated to self-correction. The agent can learn to invoke backtracking explicitly, rather than approximating it through repeated continuous cropping~\cite{yang2025longvt, pan2025timesearchr, zeng2026videoo3}.

\subsection{Multi-Round Tree Navigation}
\label{sec:inference}

At each turn, the agent materializes the children of the current node, receives an observation summarizing the current node and its children, generates a reasoning trace, and selects one of the four actions.
The loop terminates when the agent emits \texttt{answer}, reaches the maximum number of turns, or falls into a navigation cycle (Appendix A).
To support long search horizons within a fixed context budget, we maintain a compact memory comprising (i) the current state of the materialized tree, where each visited node is annotated with a short caption and a visited flag, and (ii) a chronological log of all actions taken. We use the same \model~model for both navigation and caption generation through two separate calls. The first call generates the reasoning trace and navigation action conditioned on the question, sampled frames, and memory. The second call generates short captions describing the child segments. These captions are cached, so previously processed segments do not need to be captioned again. The agent receives raw frames only for the current node, while all other nodes are represented by their corresponding textual descriptions.
This separation of detailed perception from compressed history lets the agent reason over many turns without exhausting the context budget. Detailed memory structures, caption generation procedures, and prompts are provided in Appendix A.

\subsection{Trajectory Synthesis for Hierarchical Search}
\label{sec:trajectory_synthesis}

Optimal-descent trajectories alone do not teach an agent to recognize and recover from mistakes. We therefore synthesize trajectories that combine correct hierarchical descents with deliberate wrong-branch detours followed by recovery. Given a video, question, and ground-truth evidence interval, a controller selects \texttt{zoom\_in} actions toward the highest-scoring child of the current node (scored by Qwen3-VL-8B for question relevance). When the selected child does not contain the ground-truth interval, the controller may either recover, by applying the inverse of the action sequence that led into the wrong branch, or continue further into the wrong branch to produce a longer detour. Recovery is forced after two consecutive turns in a wrong branch. Each action is paired with a reasoning trace generated by DeepSeek-R1~\cite{guo2025deepseekr1}, conditioned on the memory at that turn, the current node's frames, and the chosen action. The trajectories are derived from CG-Bench, Haystack-Ego4D, and our automatically generated LongClueQA corpus, which leverages long unlabeled YouTube videos. We focus on examples with a single ground-truth clue spanning at most $20\%$ of the video duration. Full filtering and trajectory generation details are in the Appendix.

\subsection{Training}
\label{sec:training}

\paragraph{Supervised fine-tuning.}
We first fine-tune the agent to imitate the generated trajectories. Each trajectory $\mathcal{T} = \{(O_0, M_0), (R_1, A_1, O_1, M_1), \ldots, (R_T, A_T)\}$ is decomposed into per-turn supervision examples, where at turn $t$ the input is the previous observation and current memory $(O_{t-1}, M_t)$ and the target is the reasoning trace and action $(R_t, A_t)$. Let $\mathcal{D}_{\text{SFT}}$ denote the resulting set of supervised turns and $\theta$ the model parameters. We optimize the standard token-level negative log-likelihood  $\mathcal{L}_{\text{SFT}} = -\sum_{\mathcal{D}_{\text{SFT}}} \log p_\theta(R_t, A_t \mid O_{t-1}, M_t, Q)$.
Crucially, we apply the loss selectively: although every trajectory is generated as an end-to-end sequence (including the deliberate wrong-branch detours that set up subsequent recovery turns), we backpropagate only through turns whose actions are themselves desirable as supervision --- correct \texttt{zoom\_in}, \texttt{zoom\_out} from a wrong node, and \texttt{shift} from a wrong node to the correct sibling. Turns whose actions are intentional mistakes (e.g., the \texttt{zoom\_in} into a wrong node that sets up a later recovery) remain in the model's input context as part of the trajectory history but are masked out of the loss. The agent, therefore, observes the full detour-and-recovery sequence as context but is supervised only on the moves we want it to imitate.

\paragraph{Reinforcement learning.}
We further refine the policy beyond the supervised checkpoint with reinforcement learning. Each rollout produces a complete trajectory $\tau$ that ends in an answer, and we score it with a weighted sum of three outcome-based rewards on grounding and answer correctness:
\begin{equation}
R(\tau) = \lambda_{\text{fmt}} R_{\text{fmt}}(\tau)
       + \lambda_{\text{IoU}} R_{\text{IoU}}(\tau)
       + \lambda_{\text{acc}} R_{\text{acc}}(\tau).
\end{equation}
Here, $R_{\text{fmt}}$ checks that all actions are syntactically well-formed; $R_{\text{IoU}}$ measures the temporal overlap between the predicted and ground-truth evidence intervals; $R_{\text{acc}}$ is the answer accuracy reward. We optimize $R(\tau)$ with GRPO~\cite{shao2024deepseekmath} regularized by a KL penalty against the supervised checkpoint to prevent the policy from drifting too far from the supervised initialization. The exact reward weights and the full optimization setup are described in the Appendix.

\subsection{Implementation Details}
\label{sec:implementation}

We instantiate \model~with Qwen3-VL-8B~\cite{qwen3vl} as the base model. The adaptive segmentation procedure of \S\ref{sec:tree_construction} uses $k = 1.5$. The agent samples $64$ frames per turn at $1$ fps. The trajectory synthesis pipeline yields $5{,}966$ filtered training trajectories drawn from three sources: CG-Bench~\cite{cgbench}, Haystack-Ego4D~\cite{lvhaystack}, and \textbf{LongClueQA}, a corpus of timestamped question-clue pairs that we automatically generate from unlabeled long-form YouTube videos (described in the Appendix). Supervised fine-tuning and reinforcement learning are both performed on $4{\times}$H100 GPUs. We refer to the supervised checkpoint as \model-SFT and the reinforcement-learning checkpoint as \model-RL. \model-RL is our primary model in the experiments. Full hyperparameters are reported in the appendix.

%% file: figures/fig_method.tex
\begin{figure*}[t]
    \centering
	\includegraphics[width=0.95\linewidth]{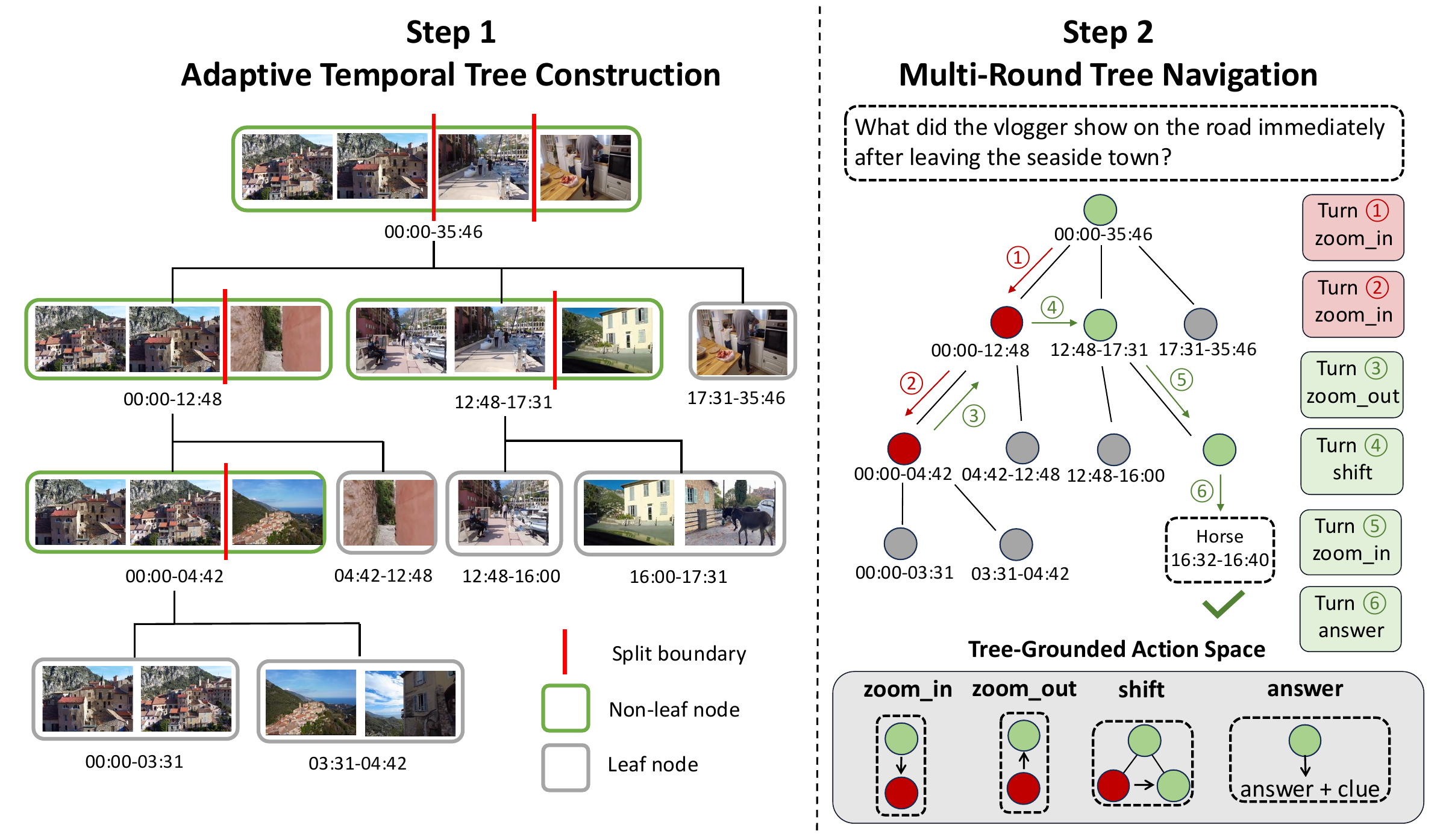}
    \caption{\textbf{Overview of VideoTreeSearch (VTS).} \textbf{Left}: Given a long video, VTS recursively partitions it into a non-uniform tree using adaptive temporal segmentation, so each node corresponds to a semantically coherent segment. \textbf{Right}: The agent navigates the tree through four discrete actions---\texttt{zoom\_in} to descend into a child, \texttt{zoom\_out} to ascend to the parent, \texttt{shift} to move laterally to a sibling, and \texttt{answer} to terminate with a final answer and evidence interval. In this example, the agent first descends into a wrong branch (red, turns 1--2), recovers via \texttt{zoom\_out} and \texttt{shift} (turns 3--4), then zooms into the correct branch (turn 5) and answers with the localized clue at 16:32--16:40 (turn 6). Backtracking is thus an explicit, learnable primitive.
    }
    \label{fig:method}
    \vspace{-0.1in}
\end{figure*}

%% file: sections/4_experiments.tex
\section{Experimental Setup}
\label{sec:setup}

\paragraph{Benchmarks and metrics.}
We evaluate \model~on three Grounded LVQA benchmarks and three general long-video QA benchmarks. For Grounded LVQA, we use the mini set of CG-Bench~\cite{cgbench} along with Haystack-LVBench~\cite{lvhaystack} and Haystack-Ego4D~\cite{lvhaystack}. We apply a filtering procedure for CG-Bench (described in the Appendix) to remove instances with unannotated or ambiguous evidence intervals. CG-Bench reports mIoU between predicted and ground-truth evidence intervals together with QA accuracy. The Haystack benchmarks report Temporal F1 (T-F1, the main grounding metric) along with QA accuracy. For general long-video QA, we evaluate on Video-MME~\cite{videomme}, the test split of MLVU~\cite{mlvu}, and LVBench~\cite{lvbench}, all measured by multiple-choice answer accuracy.

\input{tables/grounded_lvqa}

\paragraph{Baselines.}

We compare \model~against three families of methods. \emph{Uniform sampling} applies a vision-language model directly to uniformly sampled frames, with no agentic interaction. We evaluate Qwen3-VL-8B~\cite{qwen3vl}, the same base model as \model, at $256$ and $384$ frames, and Video-R1~\cite{videor1} at its optimal $64$-frame setting. \emph{Agents with proprietary LLMs} rely on proprietary or similarly large LLMs for reasoning. We compare against SiLVR~\cite{zhang2025silvr}, which uses DeepSeek-R1~\cite{guo2025deepseekr1} as the reasoner over densely extracted visual captions, DVD~\cite{zhang2025dvd}, which uses o3-mini~\cite{openai2025o3mini} to search over a multi-granular video database, and AVP~\cite{wang2025avp}, which uses Gemini-3-Flash~\cite{google2025gemini3flash} to iteratively seek evidence from the video. \emph{Multi-turn cropping agents} localize the evidence interval through continuous \texttt{crop\_video(start, end)} actions, and include methods such as LongVT~\cite{yang2025longvt}, Video-o3~\cite{zeng2026videoo3}, VideoZoomer~\cite{ding2025videozoomer}, and TimeSearch-R~\cite{pan2025timesearchr}.

\section{Experimental Results}

\subsection{Main Results on Grounded Long-Video Question Answering}
\label{sec:tg_main}

Table~\ref{tab:tg_main} reports results on the three Grounded LVQA benchmarks. \model~substantially outperforms every multi-turn cropping baseline. On CG-Bench, \model~reaches $16.8$ mIoU and $36.4$ accuracy, an absolute improvement of $+12.5$ mIoU and $+10.3$ accuracy over the strongest cropping baselines LongVT and Video-o3, respectively. On Haystack-LVBench, it reaches $15.2$ T-F1, $+7.1$ over TimeSearch-R. On Haystack-Ego4D, it reaches $18.4$ T-F1, $+7.4$ over the strongest multi-turn cropping baseline. \model~averages $4.8$ turns per video on CG-Bench (Table~\ref{tab:turn_budget} in the Appendix), whereas prior multi-turn cropping methods converge in only $1$-$2$ turns, confirming that the cropping agents commit to an interval early while \model~continues to explore and self-correct over multiple turns.
\model~also outperforms the agents with proprietary LLMs in grounding on all three benchmarks while using a smaller open-source backbone. On CG-Bench, \model~improves mIoU by $+6.7$, $+3.3$, and $+0.3$ over SiLVR, DVD, and AVP, respectively. AVP achieves higher QA accuracy than \model, which we attribute to its larger proprietary backbone (Gemini-3-Flash).

\input{tables/efficiency}

\leavevmode In Table~\ref{tab:efficiency}, we analyze the inference efficiency of \model~and compare it against Qwen3-VL-8B and four agentic methods (SiLVR, LongVT, DVD, and AVP).
We also report a more efficient variant (\model-fast), which uses at most five turns and generates each action and the corresponding captions in a single VLM call.
From the table, we can see that \model~achieves the best mIoU while processing fewer frames than all baseline methods. With a comparable runtime to SiLVR ($61.6$s vs.\ $62.3$s), \model~improves mIoU by $6.7$ and accuracy by $4.6$. Compared to DVD and AVP, \model~achieves higher mIoU while running $7.2\times$ and $1.6\times$ faster, respectively. Notably, \model-fast outperforms Qwen3-VL-8B by $1.0$ mIoU and $5.5$ accuracy with a comparable runtime ($26.5$s vs.\ $25.1$s). We also provide the runtime breakdown of \model~in Table~\ref{tab:runtime_breakdown} in the Appendix. These results show that \model~is both efficient and effective.

\subsection{Performance Gains Analysis}
\label{sec:hierarchy_analysis}

\input{tables/gains_analysis}

We attribute \model's gains to three key components: interactive search over the tree, hierarchical search across multiple temporal scales, and discrete navigation actions that make recovery from wrong branches explicit. Table~\ref{tab:ablate_combined} investigates the contributions of these components.

\noindent\textbf{Interactive search outperforms static tree representations.}
To examine the benefits of interactive tree navigation, we evaluate a static full-tree baseline inspired by VideoTree~\cite{wang2024videotree}. The baseline provides Qwen3-VL-8B with captions of all tree nodes generated by Qwen3-VL-8B itself, and predicts the answer and evidence in a single pass, without iterative search. As shown in Table~\ref{tab:ablate_combined}, \model~outperforms this baseline by $6.1$--$7.1$ points on temporal grounding metrics and $10.1$--$22.4$ points on QA accuracy across all three benchmarks, demonstrating the effectiveness of interactive tree navigation over static hierarchical representations.

\noindent\textbf{Tree-grounded actions outperform continuous cropping.}
Replacing the tree-grounded actions with a continuous \texttt{crop\_video} action under the same data and recipe degrades performance by $1.3$--$4.2$ points on temporal grounding metrics and by up to $9.8$ points on QA accuracy, indicating that the tree-grounded action space is more effective than continuous cropping.

\noindent\textbf{Self-correction is critical.}
Removing the backtracking primitives \texttt{zoom\_out} and \texttt{shift} degrades performance on every benchmark. Specifically, temporal grounding drops by $2.0$--$2.9$ points on all datasets while QA accuracy drops by $1.4$--$5.4$ points, indicating that explicit backtracking lets the agent recover from wrong descents, thereby improving both temporal grounding and answer accuracy. Additionally, we count any \texttt{zoom\_out} or \texttt{shift} in a \model~trajectory as a self-correction event, and any continuous-cropping turn whose predicted interval has IoU $<$ $0.2$ with the previous turn's interval as the analogous event. \model~self-corrects in roughly $60\%$ of trajectories on all three benchmarks, while the continuous-cropping baseline trained on identical data self-corrects in only $7$-$15\%$. Among self-correcting trajectories, $42.7\%$ subsequently reach a node containing the ground-truth evidence, indicating that the agent recovers on a substantial fraction of these harder cases.

\noindent\textbf{Hierarchical search supports a wide range of clue durations.}
Replacing the hierarchical tree with a flat tree of adaptive short segments degrades QA accuracy by $1.9$-$5.1$ points on all datasets. The flat tree achieves higher temporal grounding than the hierarchical tree only on Haystack-Ego4D ($18.9$ vs.\ $18.4$ T-F1), which has the shortest average clue duration ($9.5$s).
On the other two benchmarks, clue durations vary widely across queries, from a few seconds to minutes (mean $19.2$s on CG-Bench and $96.2$s on Haystack-LVBench), so no single flat partition matches them all. Hierarchical search lets the agent navigate to the appropriate scale for each query.
\subsection{Ablation Studies}
\label{sec:ablations}

We isolate the contribution of \model's design choices and examine how the framework generalizes across training data sources and backbones. More ablation studies are provided in the Appendix.

\input{tables/ablations}

\noindent\textbf{Tree construction.} In Table~\ref{tab:ablations_combined}, we compare our adaptive temporal segmentation with two alternative tree construction strategies: uniform splitting, which divides each segment into a fixed number of equal-length children, and OmniShotCut~\cite{wang2026omnishotcut}, a recent shot boundary detection method.
From the table, we can see that adaptive temporal segmentation outperforms uniform splitting with $4$ and $16$ children by $1.2$ and $2.9$ mIoU, and OmniShotCut by $2.8$ mIoU and $3.9$ accuracy.
These results demonstrate that adaptive temporal segmentation constructs a more effective search tree than both uniform splitting and shot boundary detection.

\noindent\textbf{Training stages.} \model~is trained in two stages, and Table~\ref{tab:ablations_combined} shows that each stage helps. The zero-shot variant, using only the action space and prompting with no training, reaches $12.8$ mIoU. SFT on synthesized trajectories adds $+2.1$ mIoU, and RL adds a further $+1.9$.

\noindent\textbf{Trajectory synthesis.} Table~\ref{tab:ablations_combined} shows that training on trajectories with deliberate detours and recovery outperforms training on optimal-path-only trajectories by $+6.4$ mIoU and $+5.1$ accuracy. This shows that the agent learns to search effectively from exposure to recovery in its training trajectories.

\input{tables/data_source_backbone}

\noindent\textbf{Effect of training data source.}
We compare training on LongClueQA alone, generated entirely from unlabeled long-form videos, against training on all sources. Note that LongClueQA has no video overlap with any evaluation benchmark. As shown in Table~\ref{tab:data_source}, \model~trained on LongClueQA alone already surpasses the strongest cropping baseline on each benchmark, by $+11.0$ mIoU on CG-Bench, $+6.4$ T-F1 on Haystack-LVBench, and $+5.8$ T-F1 on Haystack-Ego4D. Adding in-domain training videos from CG-Bench and Haystack-Ego4D (the all-sources variant) improves grounding further by $1.5$, $0.7$, and $1.6$ points, respectively. The results suggest that synthetic timestamped QA from unlabeled videos is a strong training signal on its own.

\noindent\textbf{Backbone generalization.}
\label{sec:base_model}
To check that \model's gains do not depend on the Qwen3-VL-8B backbone, we also train \model~on Qwen2.5-VL-7B~\cite{bai2025qwen25vl}, the backbone used by prior multi-turn cropping agents. Under this matched-backbone setting (Table~\ref{tab:ablate_backbone}), \model~improves over the strongest Qwen2.5-VL-7B cropping baseline by $+10.2$ mIoU and $+4.6$ accuracy on CG-Bench, $+4.2$ T-F1 and $+1.0$ accuracy on Haystack-LVBench, and $+2.1$ T-F1 and $+1.8$ accuracy on Haystack-Ego4D. The gains therefore stem from the framework rather than the base model, and a stronger backbone adds further gains.

\subsection{General Long-Video Understanding}
\label{sec:lvu_main}

\input{tables/general_lvu}

We test whether \model's tree-grounded navigation policy generalizes beyond Grounded LVQA on three general long-video QA benchmarks: Video-MME, MLVU, and LVBench. These benchmarks evaluate QA accuracy alone, with no temporal grounding, so when the agent answers, \model~pools the frames it observed across all visited nodes and conditions the answer on this pooled set rather than on a single node. Table~\ref{tab:general_lvu} reports the results. \model~outperforms every multi-turn cropping baseline on all three benchmarks: $+0.5$ on Video-MME over LongVT, $+2.4$ on MLVU over VideoZoomer, and $+7.1$ on LVBench over Video-o3. The tree-grounded navigation policy thus generalizes beyond evidence localization, despite being trained primarily for Grounded LVQA.

\begin{figure}[t!]
  \centering
  \includegraphics[width=0.9\linewidth]{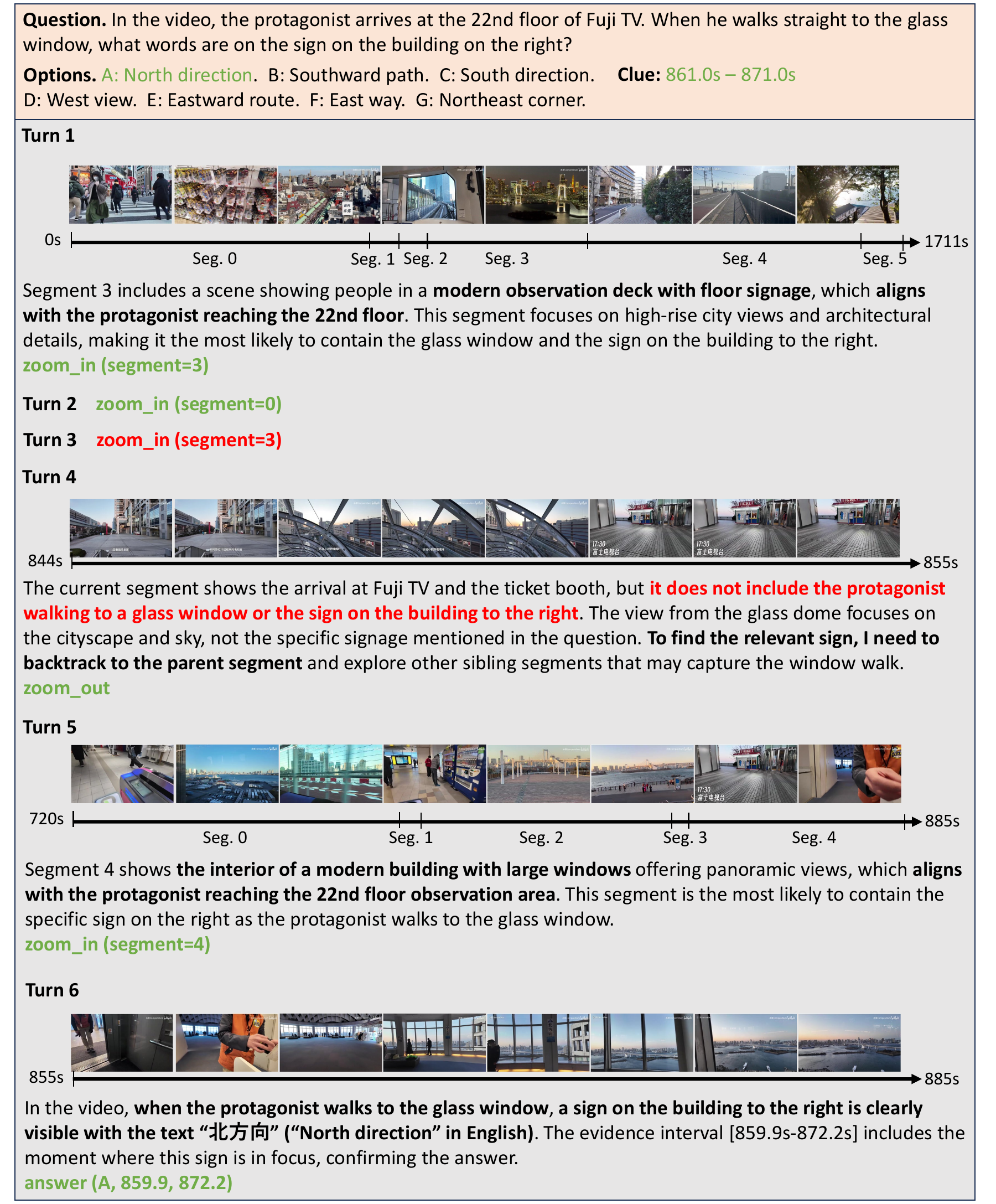}
  \vspace{-0.05in}
    \caption{\textbf{Qualitative example: recovery via \texttt{zoom\_out}.} For a question about what is visible through a 22nd-floor window of the Fuji TV building, the agent first descends into a wrong branch showing the building's ground-floor ticket booth (turns~1--3). It detects the error, issues \texttt{zoom\_out} to return to the parent (turn~4), and explores a different child to reach the correct clue (turns~5--6). The example shows \texttt{zoom\_out} as an explicit recovery action after the agent commits to a wrong branch.}
    \vspace{-0.2in}
  \label{fig:qual2}
\end{figure}

\subsection{Qualitative Analysis}
Figure~\ref{fig:qual2} shows a six-turn trajectory on a question asking what is visible through the 22nd-floor glass window. On the third turn the agent commits to a wrong branch, descending into a segment showing the building's ground-floor ticket booth. Recognizing that the ticket booth is not the queried 22nd-floor window, the agent issues \texttt{zoom\_out} to return to the parent, then re-descends through a different child to reach the correct clue. Additional qualitative examples are included in the Appendix.

%% file: tables/grounded_lvqa.tex
\begin{table}[t]
\centering
\caption{\textbf{Grounded LVQA results.} \model~substantially improves both grounding metrics (mIoU, T-F1) and QA accuracy over prior multi-turn cropping agents across all benchmarks.}
\label{tab:tg_main}
\small
\begin{tabular}{lccccccc}
\toprule
& \multicolumn{2}{c}{CG-Bench} & \multicolumn{2}{c}{Haystack-LVBench} & \multicolumn{2}{c}{Haystack-Ego4D} \\
\cmidrule(lr){2-3} \cmidrule(lr){4-5} \cmidrule(lr){6-7}
Method &  mIoU & Acc & T-F1 & Acc & T-F1 & Acc \\
\midrule
\multicolumn{7}{l}{\textit{Uniform sampling}} \\
Qwen3-VL-8B (256 frames)~\cite{qwen3vl}      & 12.1 & 20.2 & 12.8 & 45.9 & 8.9  & 39.2 \\
Qwen3-VL-8B (384 frames)~\cite{qwen3vl}      & 12.1 & 23.6 & 9.2  & 39.5 & 7.0  & 33.4 \\
Video-R1 (64 frames)~\cite{videor1}          & 5.5  & 15.8 & 4.3  & 40.6 & 2.9  & 33.1 \\
\midrule
\multicolumn{7}{l}{\textit{Agents with proprietary LLMs}} \\
SiLVR~\cite{zhang2025silvr}                  & 10.1 & 31.8 & 8.5  & 57.0 & 6.3  & 46.4 \\
DVD~\cite{zhang2025dvd}                & 13.5 & 35.0 & -- & -- & -- & -- \\
AVP~\cite{wang2025avp}                 & 16.5 & \textbf{54.2} & 10.1 & \textbf{76.1} & 16.3 & \textbf{70.0} \\
\midrule
\multicolumn{7}{l}{\textit{Multi-turn cropping agents}} \\
LongVT~\cite{yang2025longvt}                 & 4.3  & 17.4 & 5.4  & 46.2 & 3.7  & 42.1 \\
Video-o3~\cite{zeng2026videoo3}              & 2.9  & 26.1 & --   & --   & --   & --   \\
VideoZoomer~\cite{ding2025videozoomer}       & 3.0  & 18.3 & 6.8  & 48.3 & 5.4  & 48.6 \\
TimeSearch-R~\cite{pan2025timesearchr}       & --   & --   & 8.1  & 52.1 & 11.0 & 53.5 \\
\midrule
\rowcolor{gray!12}
\textbf{\model~(Ours)}                       & \textbf{16.8} & 36.4 & \textbf{15.2} & 58.3 & \textbf{18.4} & 66.1 \\
\bottomrule
\end{tabular}
\vspace{-0.1in}
\end{table}

%% file: tables/efficiency.tex
\begin{wraptable}{r}{0.46\textwidth}
\centering
\vspace{-0.1in}
\caption{\textbf{Efficiency analysis on CG-Bench.} \model~achieves the best mIoU and the best balance between performance and runtime. Runtime is measured per video on a single H100 GPU. $^\dagger$Uses a proprietary LLM.}
\label{tab:efficiency}
\footnotesize
\setlength{\tabcolsep}{2.5pt}
\begin{tabular}{lcccc}
\toprule
Method & Frames & Time (s) & mIoU & Acc \\
\midrule
 Qwen3-VL-8B~\cite{qwen3vl}   & 384 & \textbf{25.1} & 12.1 & 23.6 \\
SiLVR~\cite{zhang2025silvr} & 450 & 62.3 & 10.1 & 31.8 \\
LongVT~\cite{yang2025longvt} & 512 & 28.4 & 4.3 & 17.4 \\
DVD$^\dagger$~\cite{zhang2025dvd} & 1 fps & 442.0 & 13.5 & 35.0 \\
AVP$^\dagger$~\cite{wang2025avp} & 512 & 99.5 & 16.5 & \textbf{54.2} \\
\midrule
\model-fast (Ours) & \textbf{192} & 26.5 & 13.1 & 29.1 \\
\rowcolor{gray!12}
\model~(Ours)      & 328 & 61.6 & \textbf{16.8} & 36.4 \\
\bottomrule
\end{tabular}
\end{wraptable}

%% file: tables/gains_analysis.tex
\begin{table}[t]
\centering
\caption{\textbf{Performance gains analysis on Grounded LVQA.} Interactive search, the tree-grounded action space, the backtracking primitives, and hierarchical descent each contribute to the strong performance of \model.}
\label{tab:ablate_combined}
\small
\begin{tabular}{lcccccc}
\toprule
& \multicolumn{2}{c}{CG-Bench} & \multicolumn{2}{c}{Haystack-LVBench} & \multicolumn{2}{c}{Haystack-Ego4D} \\
\cmidrule(lr){2-3} \cmidrule(lr){4-5} \cmidrule(lr){6-7}
Variant & mIoU & Acc & T-F1 & Acc & T-F1 & Acc \\
\midrule
 Static full tree (no interactive search)               & 9.7 & 18.1 & 9.1 & 48.2 & 12.1 & 43.7 \\
Continuous \texttt{crop\_video} action       & 15.5 & 32.5 & 13.2 & 49.1 & 14.2 & 56.3 \\
Tree-grounded, no backtracking               & 14.4 & 35.0 & 13.2 & 52.9 & 15.5 & 64.6 \\
Tree-grounded, flat tree (no hierarchy)      & 15.3 & 34.5 & 13.2 & 53.2 & \textbf{18.9} & 62.9 \\
\rowcolor{gray!12}
\textbf{\model~(full design)}                & \textbf{16.8} & \textbf{36.4} & \textbf{15.2} & \textbf{58.3} & 18.4 & \textbf{66.1} \\
\bottomrule
\end{tabular}
\vspace{-0.1in}
\end{table}

%% file: tables/ablations.tex
\begin{wraptable}{r}{0.5\textwidth}
\vspace{-0.2in}
\centering
\caption{\textbf{Ablation studies on CG-Bench.} Each ablation varies one component and keeps the rest at their default setting.}
\label{tab:ablations_combined}
\footnotesize
\setlength{\tabcolsep}{3pt}
\begin{tabular}{lcc}
\toprule
Variant & mIoU & Acc \\
\midrule
\textit{Tree construction} & & \\
Uniform splitting, $4$ children                          & 15.6 & 34.9 \\
Uniform splitting, $16$ children                         & 13.9 & 33.0 \\
OmniShotCut~\cite{wang2026omnishotcut} shot boundaries & 14.0 & 32.5 \\
\rowcolor{gray!12}
Adaptive temporal segmentation (Ours)              & \textbf{16.8} & \textbf{36.4} \\
\midrule
\textit{Training stages} & & \\
Zero-shot (action space + prompting)                     & 12.8 & 20.9 \\
\model-SFT (+ synthesized trajectories)                  & 14.9 & 30.7 \\
\rowcolor{gray!12}
\model-RL (+ reinforcement learning)                     & \textbf{16.8} & \textbf{36.4} \\
\midrule
\textit{Trajectory synthesis} & & \\
Optimal-path-only trajectories                           & 10.4 & 31.3 \\
\rowcolor{gray!12}
Detour-and-recovery trajectories (default)               & \textbf{16.8} & \textbf{36.4} \\
\bottomrule
\end{tabular}
\vspace{-0.2in}
\end{wraptable}

%% file: tables/data_source_backbone.tex
\begin{table}[t]
\centering
\caption{\textbf{Effect of training data source and backbone.} We compare training \model~on LongClueQA alone, our corpus of synthetic timestamped QA from unlabeled long-form videos, against training on all sources, which adds in-domain videos from CG-Bench and Haystack-Ego4D. We also train \model~on Qwen2.5-VL-7B, the backbone used by prior multi-turn cropping agents. The reference row reports the strongest prior multi-turn cropping baseline per metric (LongVT for CG-Bench mIoU, Video-o3 for CG-Bench accuracy, TimeSearch-R for Haystack-LVBench and Haystack-Ego4D).}
\label{tab:data_source}
\label{tab:ablate_backbone}
\small
\setlength{\tabcolsep}{2pt}
\begin{tabular}{llcccccc}
\toprule
& & \multicolumn{2}{c}{CG-Bench} & \multicolumn{2}{c}{Haystack-LVBench} & \multicolumn{2}{c}{Haystack-Ego4D} \\
\cmidrule(lr){3-4} \cmidrule(lr){5-6} \cmidrule(lr){7-8}
Training data source & Backbone & mIoU & Acc & T-F1 & Acc & T-F1 & Acc \\
\midrule
LongClueQA only       & Qwen3-VL-8B & 15.3 & 28.7 & 14.5 & 57.5 & 16.8 & 53.7 \\
All sources     & Qwen2.5-VL-7B & 14.5 & 30.7 & 12.3 & 53.1 & 13.1 & 55.3 \\
All sources           & Qwen3-VL-8B & \textbf{16.8} & \textbf{36.4} & \textbf{15.2} & \textbf{58.3} & \textbf{18.4} & \textbf{66.1} \\
\midrule
\textit{Reference: strongest cropping baseline} & Qwen2.5-VL-7B & 4.3 & 26.1 & 8.1 & 52.1 & 11.0 & 53.5 \\
\bottomrule
\end{tabular}
\vspace{-0.2in}
\end{table}

%% file: tables/general_lvu.tex
\begin{wraptable}{r}{0.47\textwidth}
\vspace{-0.2in}
\centering
\caption{\textbf{General long-video understanding.} Accuracy (\%) on Video-MME, MLVU, and LVBench, with frame counts for \model~averaged across the three. All MLVU results are on the test split. $^*$Reproduced on the MLVU test split using the official Video-o3 checkpoint.}
\label{tab:general_lvu}
\footnotesize
\setlength{\tabcolsep}{2.5pt}
\begin{tabular}{lcccc}
\toprule
Method & Frames & V-MME & MLVU & LVB \\
\midrule
LongVT~\cite{yang2025longvt}                       & 512--768  & 67.0 & --   & 41.3 \\
Video-o3~\cite{zeng2026videoo3}                    & $\leq$768 & 66.5 & 53.1$^*$ & 47.6 \\
TimeSearch-R~\cite{pan2025timesearchr}             & 768       & 66.6 & --   & --   \\
VideoZoomer~\cite{ding2025videozoomer}             & 128       & 65.2 & 55.8 & 41.5 \\
\midrule
\rowcolor{gray!12}
\textbf{\model~(Ours)}                             & 274 & \textbf{67.5} & \textbf{58.2} & \textbf{54.7} \\
\bottomrule
\vspace{-0.3in}
\end{tabular}
\end{wraptable}

%% file: sections/5_conclusion.tex
\section{Conclusion}
We presented \model, a tree-grounded agentic framework for grounded long-video question-answering. \model~organizes the video as a tree of semantically coherent segments and navigates it with discrete actions, which turns precise localization and backtracking from implicit capabilities into explicitly learnable behaviors. On three grounded long-video QA benchmarks, \model~substantially outperforms prior multi-turn cropping agents and achieves better grounding than agents with proprietary LLMs while processing fewer frames, and it transfers to three general long-video understanding benchmarks. More broadly, our results suggest that structuring the search over a long video is a useful inductive bias for long-horizon video reasoning.

%% file: sections/7_acknowledgements.tex
\section*{Acknowledgements}
This work was supported by Laboratory for Analytic Sciences via NC State University, ONR Award N00014-23-1-2356, Sony Focused Research award, and NSF CAREER Award 2541848.

%% file: sections/8_supp.tex
\section*{Appendix}
Our appendix consists of
Additional Implementation Details (Section~\ref{sec:addtional_implementation_details}),
Evaluation Benchmarks (Section~\ref{sec:addtional_evaluation_details}),
Data Construction (Section~\ref{sec:data_construction}),
Additional Analyses (Section~\ref{sec:additional_analyses}),
More Qualitative Results (Section~\ref{sec:visualization}), and Limitations and Future Work (Section~\ref{sec:limitations}).

\section{Additional Implementation Details}
\label{sec:addtional_implementation_details}

\paragraph{Training Details.}
We use full fine-tuning of the LLM for SFT and LoRA fine-tuning for RL. SFT is conducted with the LLaMA-Factory framework and RL is conducted with the ms-swift framework. For SFT, we set the learning rate to \texttt{1e-5}, the batch size to $16$, and train for $2$ epochs. For RL, we set the learning rate to \texttt{5e-5}, the batch size to $96$, and train for $1$ epoch with GRPO using a group of $8$ trajectories per prompt. For each video segment, we sample frames at $1$ fps with a maximum of $64$ frames, and each frame is constrained to a maximum resolution of $224 \times 224$ pixels. Both SFT and RL are conducted on $4 \times$ H100 GPUs, taking approximately $4$ and $32$ hours, respectively. Full hyperparameters are reported in Table~\ref{tab:keyhyperpara}.

\begin{table}[h]
\caption{\textbf{Key training hyperparameters} for the SFT and RL stages of \model.}
\label{tab:keyhyperpara}
\begin{subtable}[t]{0.48\textwidth}
\caption{SFT stage}
\centering
\begin{tabular}{@{}ll@{}}
\toprule
\textbf{Hyperparameter} & \textbf{Value} \\
\midrule
Train epochs & \texttt{2} \\
Train batch size & \texttt{16} \\
Learning rate & \texttt{1e-5} \\
Learning rate scheduler & \texttt{cosine} \\
Warmup ratio & \texttt{0.1} \\
Freeze vision encoder & \texttt{true} \\
\bottomrule
\end{tabular}
\end{subtable}
\hfill
\begin{subtable}[t]{0.48\textwidth}
\caption{RL stage}
\centering
\begin{tabular}{@{}ll@{}}
\toprule
\textbf{Hyperparameter} & \textbf{Value} \\
\midrule
Format reward weight & \texttt{0.5} \\
Accuracy reward weight & \texttt{0.5} \\
IoU reward weight & \texttt{1.0} \\
Max interaction turns & \texttt{10} \\
LoRA rank & \texttt{32} \\
LoRA alpha & \texttt{64} \\
LoRA dropout & \texttt{0} \\
Train batch size & \texttt{96} \\
Rollout temperature & \texttt{1.0} \\
Rollouts per prompt ($n$) & \texttt{8} \\
KL coefficient ($\beta$) & \texttt{0.04} \\
Learning rate & \texttt{5e-5} \\
Learning rate scheduler & \texttt{constant} \\
\bottomrule
\end{tabular}
\end{subtable}
\end{table}

\paragraph{GRPO Objective.}
For each training question, we sample a group of $G = 8$ trajectories $\{\tau_i\}_{i=1}^{G}$ from the old policy $\pi_{\theta_{\text{old}}}$, each with at most $10$ turns. Each trajectory receives the reward $R_i = 0.5\, R_{\text{fmt}}(\tau_i) + 1.0\, R_{\text{IoU}}(\tau_i) + 0.5\, R_{\text{acc}}(\tau_i)$, and its advantage is computed within the group as $\hat{A}_i = (R_i - \text{mean}(R_{1:G})) / \text{std}(R_{1:G})$. Let $a_{i,t}$ denote the $t$-th token generated in $\tau_i$, $s_{i,t}$ its context, and $\rho_{i,t} = \pi_\theta(a_{i,t} \mid s_{i,t}) / \pi_{\theta_{\text{old}}}(a_{i,t} \mid s_{i,t})$ the importance ratio. The loss is
\begin{equation}
\begin{aligned}
\mathcal{L}_{\text{GRPO}} = -\frac{1}{G} \sum_{i=1}^{G} \frac{1}{|\tau_i|} \sum_{t=1}^{|\tau_i|} \Big[ & \min\big( \rho_{i,t} \hat{A}_i,\ \text{clip}(\rho_{i,t}, 1 - \varepsilon, 1 + \varepsilon)\, \hat{A}_i \big) \\
& - \beta\, D_{\text{KL}}\big( \pi_\theta(\cdot \mid s_{i,t}) \,\Vert\, \pi_{\text{ref}}(\cdot \mid s_{i,t}) \big) \Big],
\end{aligned}
\end{equation}
where $\pi_{\text{ref}}$ is the SFT policy, $\varepsilon = 0.2$, and $\beta = 0.04$.

\paragraph{Training Data.}
The training data is sourced from the CG-Bench full set (excluding questions from the mini set), the Haystack-Ego4D training set, and LongClueQA. In total, we have $5{,}966$ QA pairs, and $5{,}966$ trajectories from all data sources ($1{,}144$ from CG-Bench, $1{,}130$ from Haystack-Ego4D, and $3{,}692$ from LongClueQA). We randomly pick 40\% of trajectories for SFT training and 60\% of the QA pairs for RL training.

\paragraph{Adaptive Temporal Segmentation Algorithm.}
At each node of the search tree, we partition the current segment into child segments aligned with visual and semantic transitions. Algorithm~\ref{algo:adaptive_splitting} summarizes the procedure. We uniformly sample frames from the current segment, encode each with a CLIP image encoder, and compute the cosine distance between consecutive frame embeddings to obtain a per-position boundary signal $\delta$. We then form an adaptive threshold $\tau = \text{mean}(\delta) + k \cdot \text{std}(\delta)$ from the local statistics of $\delta$, and treat positions with $\delta_i > \tau$ as boundaries. We clip the boundary set so that the number of children falls within $[N_\text{min}, N_\text{max}]$: if the threshold yields too many boundaries we keep those with the largest $\delta_i$, and if it yields too few we add the next-largest until the minimum is met. We set $N_\text{min} = 3$ and $N_\text{max} = 8$. For each selected transition $\delta_i$, we place the boundary at the timestamp of the later frame $i{+}1$.

\begin{algorithm}[h]
\begin{algorithmic}[1]
\Require Segment $S = [t_\text{start}, t_\text{end}]$, CLIP encoder $E$, threshold coefficient $k$, min/max children $N_\text{min}, N_\text{max}$
\State $F \gets \texttt{sampleFrames}(S)$ \Comment{$\leq 64$ frames at $1$~fps}
\State $\{e_i\}_{i=1}^{n} \gets \texttt{encode}(F, E)$
\State $\delta_i \gets 1 - \cos(e_i, e_{i+1})$, \ for $i = 1, \ldots, n-1$
\State $\tau \gets \text{mean}(\delta) + k \cdot \text{std}(\delta)$
\State $B \gets \{\, i : \delta_i > \tau \,\}$
\If{$|B| + 1 > N_\text{max}$}
    \State $B \gets$ indices of the top $(N_\text{max} - 1)$ values in $\delta$
\ElsIf{$|B| + 1 < N_\text{min}$}
    \State $B \gets$ indices of the top $(N_\text{min} - 1)$ values in $\delta$
\EndIf
\State \Return \texttt{splitByBoundaries}($S, B$) \Comment{cut $S$ at the timestamp of frame $i{+}1$ for each $i \in B$}
\end{algorithmic}
\caption{Adaptive Temporal Segmentation}
\label{algo:adaptive_splitting}
\end{algorithm}

\paragraph{Hierarchical Tree Search Algorithm.}
Algorithm~\ref{algo:search} describes the inference-time hierarchical search. The agent starts at the root segment covering the full video and maintains a memory $\mathcal{M}$ containing (i) the partial search tree expanded so far, with a caption for each visited node, and (ii) the interaction history of prior actions and their outcomes. At each turn, the agent splits the current segment into children via Algorithm~\ref{algo:adaptive_splitting} (or retrieves them from the cached tree), is shown frames sampled from those children together with $\mathcal{M}$, and outputs an action in $\{\texttt{zoom\_in}, \texttt{zoom\_out}, \texttt{shift}, \texttt{answer}\}$. \texttt{zoom\_in} descends into a selected child, \texttt{zoom\_out} returns to the parent, and \texttt{shift} moves to a sibling under the same parent. After the action is executed, the same model $\pi$ generates a caption for each child segment in a separate call, and these captions are written into $\mathcal{M}$ together with the reasoning trace and the action. The loop terminates when the agent emits \texttt{answer}, returning the predicted answer $\hat{y}$ and evidence interval $[t_s, t_e]$. In practice, the algorithm also terminates when a navigation cycle is detected, i.e., when the agent returns to a previously visited node and selects the same action that it took at its previous visit to that node. Revisiting a node alone does not count as a cycle, so backtracking through \texttt{zoom\_out} and \texttt{shift} is unaffected. When a cycle is detected, the agent answers via \texttt{forceAnswer} using the current memory.

\begin{algorithm}[h]
\begin{algorithmic}[1]
\Require Video $V$, question $Q$, agent policy $\pi$, captioning prompt $P_\text{cap}$, max turns $T_\text{max}$
\State $S \gets [0, \text{len}(V)]$ \Comment{current node}
\State $\mathcal{M} \gets \texttt{initMemory}()$ \Comment{tree structure + interaction history}
\For{$t = 1, \ldots, T_\text{max}$}
    \State $\{S_1, \ldots, S_m\} \gets \texttt{AdaptiveTemporalSegmentation}(S)$ \Comment{Algorithm~\ref{algo:adaptive_splitting}}
    \State $O \gets \texttt{sampleFrames}(\{S_1, \ldots, S_m\})$
    \State $(R, A) \gets \pi(O, \mathcal{M}, Q)$ \Comment{reasoning and action}
    \If{$A \neq \texttt{answer}$ and $A$ was taken at $S$ in a previous turn}
        \State \Return \texttt{forceAnswer}($O$, $\mathcal{M}$, $Q$) \Comment{navigation cycle}
    \EndIf
    \If{$A = \texttt{zoom\_in}(j)$}
        \State $S \gets S_j$
    \ElsIf{$A = \texttt{zoom\_out}$}
        \State $S \gets \texttt{parent}(S)$
    \ElsIf{$A = \texttt{shift}(j)$}
        \State $S \gets$ sibling of $S$ indexed by $j$
    \ElsIf{$A = \texttt{answer}(\hat{y}, [t_s, t_e])$}
        \State \Return $\hat{y}, [t_s, t_e]$
    \EndIf
    \State $C \gets \pi(O, P_\text{cap})$ \Comment{caption each child segment with $\pi$}
    \State $\mathcal{M} \gets \texttt{updateMemory}(\mathcal{M}, C, R, A, S)$
\EndFor
\State \Return \texttt{forceAnswer}($O$, $\mathcal{M}, Q$) \Comment{if turn budget exhausted}
\end{algorithmic}
\caption{Hierarchical Tree Search}
\label{algo:search}
\end{algorithm}

\paragraph{Prompt Structure.}
We show the prompt structure used at every decision turn in the block below. The system prompt establishes the agent's role and enumerates the four discrete actions, while the per-turn user prompt provides frames from the current node's children, a textual memory block (tree structure, interaction history, and navigation context), and the question with its multiple-choice options.

\begin{tcolorbox}[
    colback=gray!10!white,
    colframe=black,
    boxrule=1pt,
    boxsep=4pt,
    top=4pt, bottom=4pt,
    left=8pt, right=8pt
]
\begin{center}
\textbf{VTS Agent Prompt}
\end{center}
\small
\textbf{System Prompt:} You are a video question-answering agent that navigates a long video through hierarchical temporal search. At each step, the current video segment is divided into non-overlapping child segments. You observe frames from each child segment and decide one of four actions:\\[3pt]
1. \texttt{zoom\_in <segment\_id>}: Examine a child segment at finer granularity.\\
2. \texttt{zoom\_out}: Backtrack to the parent segment to explore a different region.\\
3. \texttt{shift <segment\_id>}: Move to a sibling segment under the same parent.\\
4. \texttt{answer <letter> <evidence\_start> <evidence\_end>}: Provide the answer to the question along with the time interval (in seconds) that contains the supporting evidence.\\[5pt]
\textbf{User:}\\[1pt]
\texttt{Current Segment [$t_s$-$t_e$]:}\\
\quad $\dots$\\[3pt]
\quad \texttt{Segment $i$ [$t^i_s$-$t^i_e$]:} \texttt{<frame\_timestamp><image>} $\dots$\\
\quad $\dots$\\[3pt]
\\
\texttt{Memory:}\\
\quad \texttt{Tree Structure:}\\
\quad \texttt{<materialized tree with [visited]/[current] flags and short node captions>}\\
\\
\quad \texttt{Interaction History:}\\
\quad \texttt{<chronological log of past turns and actions>}\\
\\
\quad \texttt{Navigation Context:}\\
\texttt{<current position at the tree, parent, valid zoom\_in children, valid shift siblings>}\\
\\[3pt]
\texttt{Question:} $\dots$\\
\texttt{Choices:} \texttt{A.} $\dots$ \quad \texttt{B.} $\dots$ \quad \texttt{C.} $\dots$ \quad $\dots$
\end{tcolorbox}

\paragraph{Memory Structure.}
The memory is a textual block serialized into the prompt at every turn. It contains three components. The \emph{tree structure} records the materialized part of the tree, where each node contains its temporal interval, a short caption, and a status flag (not visited, visited, or current). The \emph{interaction history} records the actions taken in previous turns. The \emph{navigation context} lists the current node, its parent, and the valid \texttt{zoom\_in} and \texttt{shift} targets, which prevents invalid actions. An example of the memory at an intermediate turn is shown below.

\begin{tcolorbox}[
    colback=gray!10!white,
    colframe=black,
    boxrule=1pt,
    boxsep=4pt,
    top=4pt, bottom=4pt,
    left=8pt, right=8pt
]
\begin{center}
\textbf{Memory Example}
\end{center}
\small\ttfamily
Tree Structure:\\
{[}0s-1531s{]} {[}visited{]}, Caption: {[}...{]}\\
|-- Segment 0: {[}0s-12s{]} {[}not visited{]}, Caption: {[}...{]}\\
|-- Segment 1: {[}12s-328s{]} {[}visited{]}, Caption: {[}...{]}\\
|\quad\ \ |-- Segment 1.0: {[}12s-20s{]} {[}not visited{]}, Caption: {[}...{]}\\
|\quad\ \ |-- ...\\
|\quad\ \ `-- Segment 1.3: {[}90s-230s{]} {[}visited{]}, Caption: {[}...{]}\\
|\quad\ \ \quad\ \ |-- Segment 1.3.0: {[}90s-113s{]} {[}visited{]} {[}current{]}, Caption: {[}...{]}\\
|\quad\ \ \quad\ \ |-- Segment 1.3.1: {[}113s-149s{]} {[}not visited{]}, Caption: {[}...{]}\\
|\quad\ \ \quad\ \ `-- ...\\
`-- ...\\[4pt]
Interaction History:\\
- Turn 1: ZOOM\_IN(segment=1, {[}12s-328s{]})\\
- Turn 2: ZOOM\_IN(segment=3, {[}90s-230s{]})\\
- Turn 3: ZOOM\_IN(segment=0, {[}90s-113s{]})\\[4pt]
Navigation Context:\\
- Current Position: {[}90s-113s{]}, Parent: {[}90s-230s{]}\\
- Valid ZOOM\_IN targets: {[}...{]}\\
- Valid SHIFT targets: Segment 1.3.1 {[}113s-149s{]}, Segment 1.3.2 {[}149s-153s{]}, Segment 1.3.3 {[}153s-225s{]}, Segment 1.3.4 {[}225s-230s{]}
\end{tcolorbox}

\paragraph{Caption Generation.}
The captions in the memory are generated by \model-8B itself. At each turn, the model is called twice. The first call takes the frames, the question, and the memory as input and generates the reasoning and the action. The second call generates a short caption for each child segment. The captions are cached in the memory, so each segment is captioned only once.

\begin{figure}[h]
  \centering
  \includegraphics[width=\linewidth]{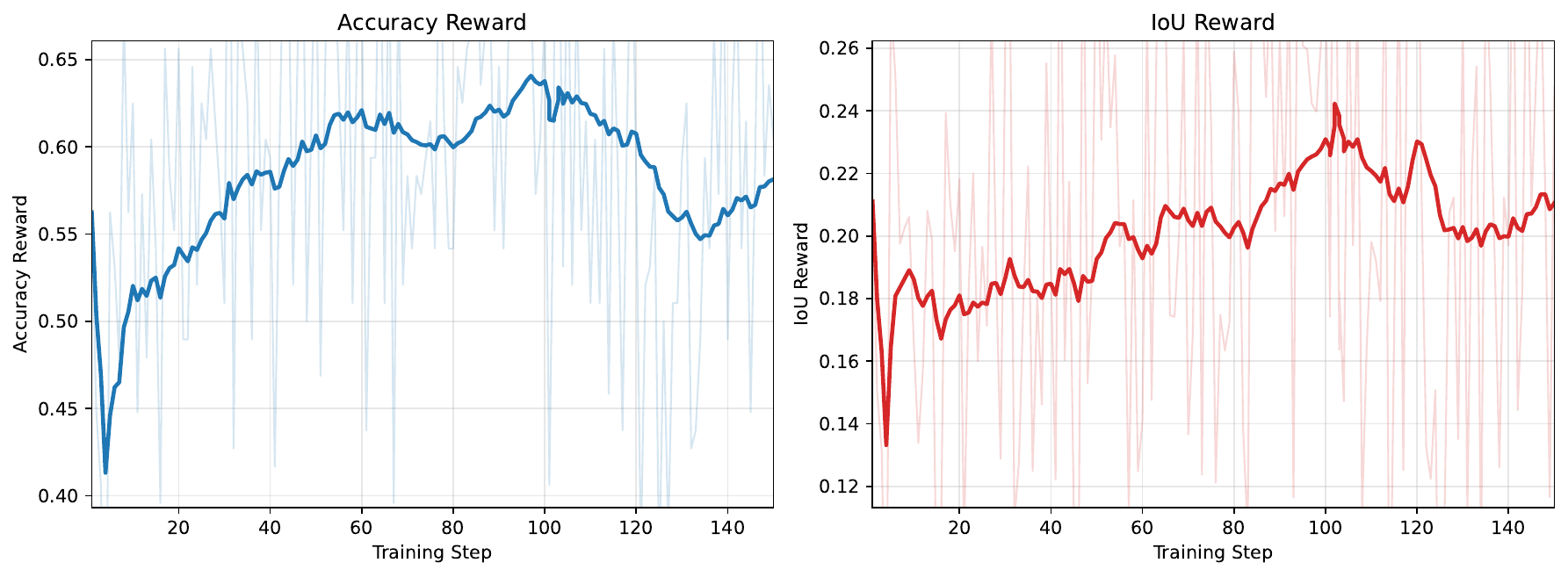}
  \caption{Accuracy (left) and temporal-IoU (right) rewards over the first 150
    GRPO updates of our best model variant.}
  \label{fig:rewards}
\end{figure}

\paragraph{Training Dynamics.}
Figure~\ref{fig:rewards} shows the training dynamics of our best GRPO model over the first 150 updates. Both the accuracy reward and temporal-IoU reward improve over training. The IoU reward is noticeably noisier, suggesting that temporal grounding is harder to optimize.

\paragraph{Inference Configuration.}
We employ the vLLM framework for inference, setting the temperature to $0$ and using greedy decoding to obtain deterministic results. The maximum number of interaction turns is set to $10$, matching the training setting. We use the official evaluation code for all three Grounded LVQA datasets: CG-Bench, Haystack-LVBench, and Haystack-Ego4D. Haystack-LVBench and Haystack-Ego4D evaluate predictions as a set of relevant frame indices rather than a temporal interval. To adapt \model's segment-level predictions to this frame-level evaluation, we uniformly sample $8$ frames from the predicted segment as the frame-level prediction.

\section{Evaluation Benchmarks}
\label{sec:addtional_evaluation_details}

\noindent \textbf{CG-Bench-mini (filtered).} CG-Bench provides multiple-choice questions paired with annotated clue intervals. The mini split contains 3,000 questions over videos averaging 28.5~min. In practice, we identify three issues with the original mini split: (1) some questions are answerable without video input; (2) the annotated clue is not always the only evidence, since the information needed to answer the question often appears outside the annotated segment as well, so the question can be answered without localizing that clue; and (3) some clue annotations are inaccurate. We address these with a four-stage filtering pipeline applied in the order below, where after each stage only the surviving questions are passed to the next.
\emph{Filter 1 (Rule-based)} applies filters that require no model inference: (i)~\emph{multi-segment removal}, dropping questions whose GT consists of more than one disjoint clue segment, since our hierarchical search formulation assumes a single contiguous target interval; (ii)~\emph{high-coverage removal}, dropping questions for which the GT segment covers more than $20\%$ of the full video duration, because such questions admit trivial trajectories and provide little supervision for temporal search; and (iii)~\emph{validity check}, removing examples with non-positive video duration or missing clue intervals. \emph{Filter 2 (Video dependency)} removes questions answerable from the question text alone: we prompt the VLM with only the question text and the multiple-choice options (no frames) and run direct inference, discarding the question if the predicted answer matches the ground-truth letter. \emph{Filter 3 (Clue sufficiency)} verifies that the annotated GT clue is informative enough to answer the question: we sample frames at $1$ fps from the GT segment (capped at $256$ frames per question), feed them to the VLM with the question and choices, and drop questions for which the VLM fails to predict the correct answer. \emph{Filter 4 (Clue uniqueness)} ensures the GT clue is the unique evidence required: we sample $1$ fps frames from the entire video excluding the GT segment (capped at $256$ frames) and prompt the VLM with these out-of-clue frames, discarding the question if the predicted answer matches the ground-truth letter, since such questions provide weak supervision for temporal grounding.
All VLM-based filters use Qwen3-VL-8B-Instruct served via vLLM with temperature $0$. The final filtered set contains 1,176 questions, averaging 28.7~min per video with a mean clue duration of 19.2~s, roughly 1.2\% of the full video.

\noindent \textbf{Haystack-LVBench} is built on LongVideoBench. We use the official 342-question validation split over 114 videos, averaging 22.9~min with 1.84 keyframes per question.

\noindent \textbf{Haystack-Ego4D} is derived from Ego4D. We use the validation split, which contains 1{,}000 questions over 71 videos, averaging 26.2~min with 2.17 keyframes per question.
\section{Data Construction}
\label{sec:data_construction}

\subsection{LongClueQA Dataset}
To train our model, we construct LongClueQA, a temporally grounded multiple-choice QA dataset constructed from unlabeled long YouTube videos. The dataset serves as a training resource for evidence-seeking behavior required by VTS: the model must not only answer a question, but also identify the short temporal interval that supports the answer.

\paragraph{Data Collection.}
\label{sec:data_collection}

We firstly collect videos from YouTube for a large and diverse pool of real-world long-form videos. Specifically, we collect candidate videos from YT-Temporal-1B~\cite{zellers2022merlotreserve}, a large-scale corpus of YouTube videos. We then perform a filtering and select videos that are between 10 to 90 minutes in length with content types such as instructional, narrative, procedural, sports, and vlog-style videos that include sufficient metadata. We also only select videos that have Creative Commons or fair-use licensing requirements. In addition, we only select videos that do not appear in CG-Bench~\cite{cgbench}, Ego4D~\cite{grauman2022ego4d}, and LongVideoBench~\cite{longvideobench}, so that there is no video overlap between LongClueQA and the evaluation benchmarks.

To ensure that the final videos contain meaningful temporal structure, we apply a visual diversity filter based on CLIP embeddings. For each candidate video, we sparsely sample frames and compute normalized CLIP features. We then measure the average pairwise cosine similarity between sampled frames. We filter out videos with low diversity scores as such videos are more likely to contain static camera setups or repeated visual content, and are therefore less useful for training temporal grounding models.

\paragraph{QA Generation.}
\label{sec:qa_generation}

We generate timestamp-anchored multiple-choice questions through a multi-stage pipeline. First, each video is segmented into 10-second clips. We then use Qwen3-VL-30B-A3B-Instruct to produce a free-form caption describing the visual content of that temporal segment. The caption is augmented with its start and end timestamp. We then use Qwen3-Next-80B-A3B-Thinking~\cite{yang2025qwen3} as question generation model and input the previously extracted captions prompted to create three questions for each video with an answer and corresponding timestamp as the supporting clue. We then generate four distractors in a separate pass with the question, answer, and ±30-second caption around the supporting clue as input. The distractors are required to be plausible, semantically similar to the correct answer, and unambiguously incorrect with respect to the video. This yields a five-way multiple-choice question that challenges model to distinguish the correct answer from plausible alternatives. In total, LongClueQA contains $3{,}692$ questions from $1{,}636$ unique videos.

\begin{figure*}[t]
    \centering
	\includegraphics[width=0.95\linewidth]{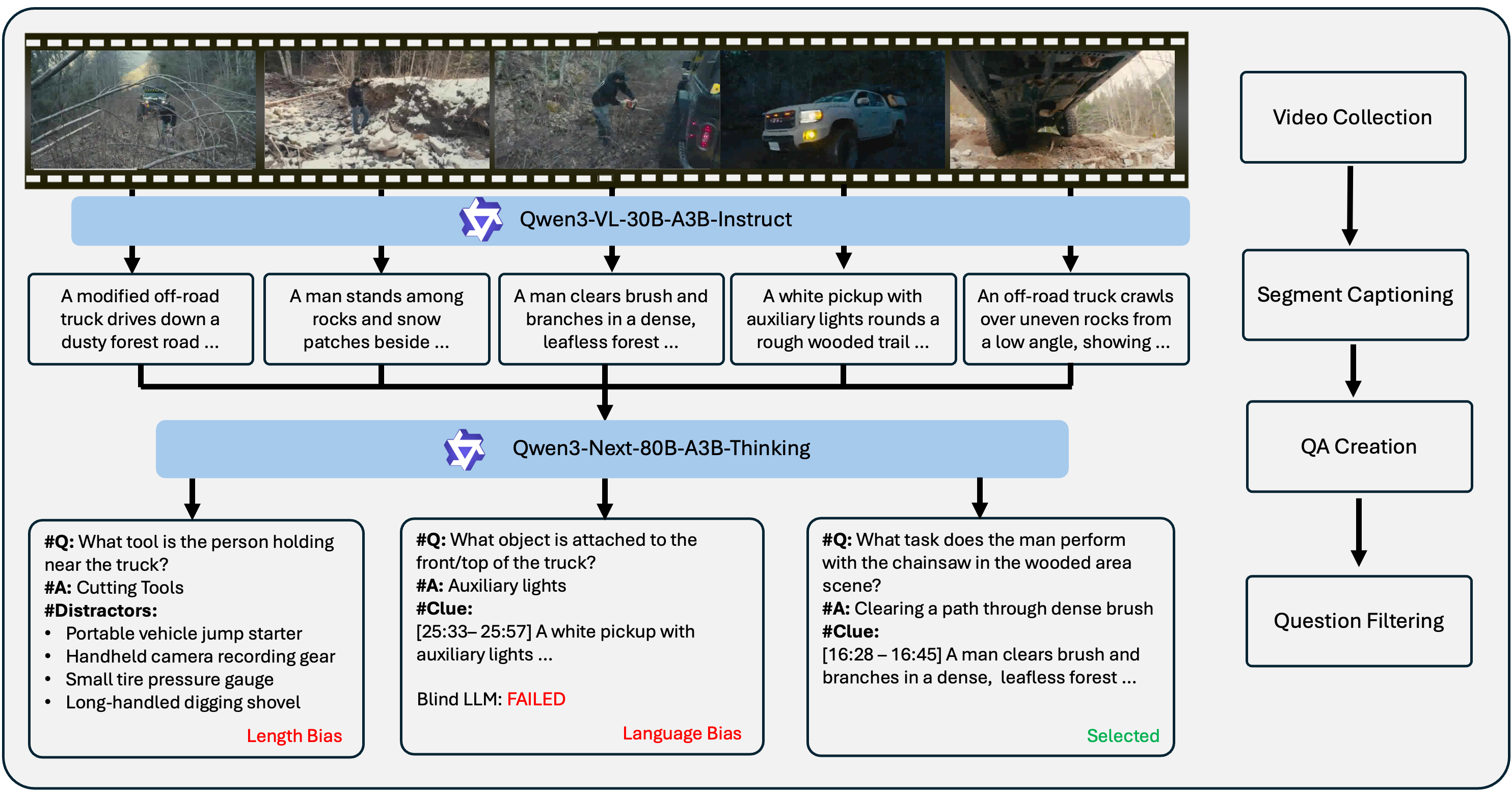}
    \caption{\textbf{Overview of the LongClueQA construction pipeline.} Collected YouTube videos are segmented into 10-second clips and captioned with Qwen3-VL. Multiple-choice QA pairs with temporal clues are then generated from the captions and filtered to retain only high-quality questions.}
    \label{fig:dataset}
\end{figure*}

\subsection{Trajectory Synthesis}
\label{sec:trajectory}

\paragraph{Data Filtering.}
We use three data sources for generating trajectories: the CG-Bench full set (excluding questions from the mini set), the Haystack-Ego4D training set, and LongClueQA. CG-Bench and LongClueQA provide ground-truth (GT) clue-segment annotations. The annotations of Haystack-Ego4D are GT frame indices. We convert these to a single segment by taking the minimum and maximum of the annotated frame timestamps. We then apply a four-stage filtering pipeline to all data sources as described in Section~\ref{sec:addtional_evaluation_details}.

\paragraph{Trajectory Generation.}
Given a long video, a question, and the ground-truth evidence interval $[t_s^{\text{GT}}, t_e^{\text{GT}}]$, the controller produces a navigation trajectory through the tree. At any point during this process, the trajectory is in one of two states: a \emph{correct} state, in which the current node overlaps the ground-truth interval, or a \emph{wrong} state, in which it does not. The current state determines how the next action is chosen. In the correct state, we score each child of the current node for question relevance using an external scorer (Qwen3-VL-8B in our main experiments) and select the highest-scoring child as the next \texttt{zoom\_in} target. If this child also contains the ground-truth interval, the trajectory remains in the correct state; otherwise it transitions into the wrong state. In the wrong state, we randomly choose between two options: (i) \emph{recover}, in which we apply the inverse of the action sequence that led into the wrong branch (e.g., a \texttt{zoom\_in} into a wrong sibling is recovered by \texttt{zoom\_out} followed by an optional \texttt{shift} to the correct sibling), or (ii) \emph{continue}, in which we deliberately take another step deeper into the wrong branch to produce a longer detour. To prevent unbounded detours, we force recovery after the trajectory has spent two consecutive turns in the wrong state. The procedure terminates when the model can produce a correct answer with sufficiently accurate evidence (mIoU $> 0.3$) at the current node, or when further descent is no longer possible.

\begin{figure}[h]
  \centering
  \includegraphics[width=\linewidth]{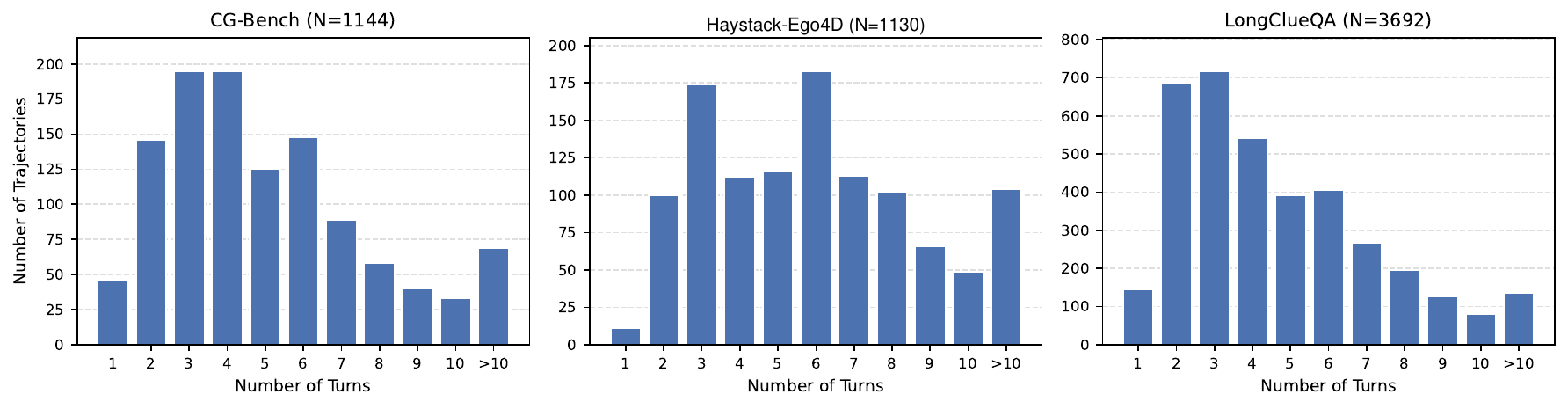}
  \caption{\textbf{Distribution of trajectory lengths across the three data sources.} For each dataset we plot the number of trajectories that terminate after a given number of turns, where one turn corresponds to a single agent action (\textsc{Zoom\_In}, \textsc{Zoom\_Out}, \textsc{Shift}, or \textsc{Answer}). Trajectories with more than $15$ turns are discarded as outliers, and lengths beyond $10$ are aggregated into a single ``$>10$'' bin.}
  \label{fig:turn-dist}
\end{figure}

\paragraph{Trajectory Statistics}
Figure~\ref{fig:turn-dist} reports the distribution of trajectory lengths across the three data sources. CG-Bench and LongClueQA trajectories are typically short (median $4$ turns), while Haystack-Ego4D requires deeper search (median $6$). The gap reflects the difficulty of each dataset's temporal-grounding queries: shorter trajectories correspond to questions answerable after a few coarse zooms, while longer ones require deeper traversal of the segment tree.

\begin{figure}[h]
  \centering
  \includegraphics[width=\linewidth]{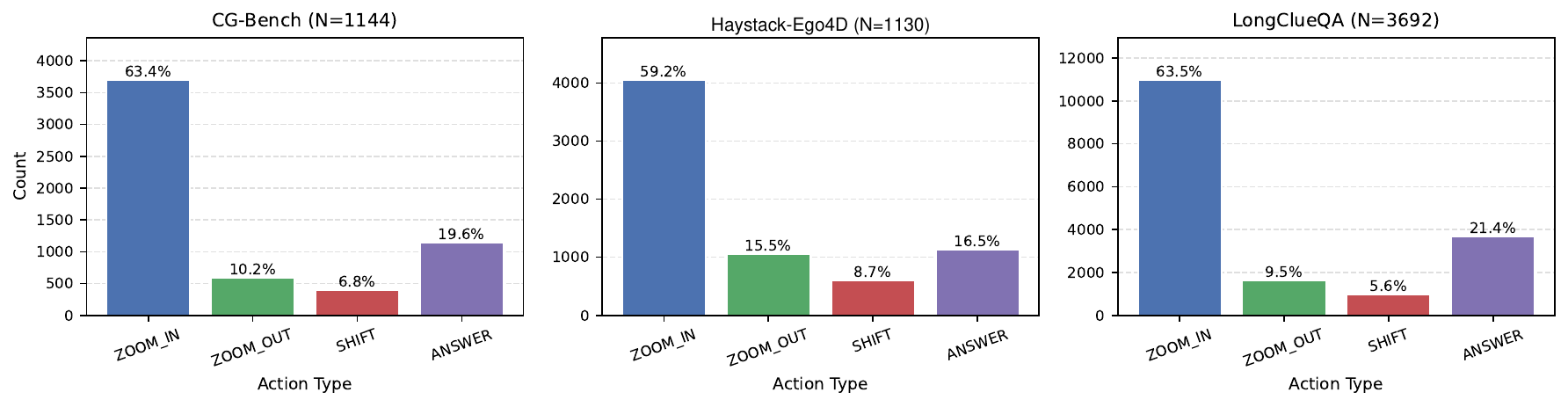}
  \caption{\textbf{Action type distribution across the three data sources.} Counts of the four action types issued by the agent across all trajectories from the three data sources.}
  \label{fig:action-dist}
\end{figure}

Figure~\ref{fig:action-dist} shows the breakdown of action types issued during trajectory generation. \textsc{Zoom\_In} dominates on every dataset, reflecting the top-down nature of hierarchical search. \textsc{Answer} appears once per trajectory and therefore tracks dataset size. \textsc{Zoom\_Out} and \textsc{Shift} act as corrective moves, and their relative frequency indicates how often the initial descent path needs to be revised before the agent commits to an answer.

\section{Additional Analyses}
\label{sec:additional_analyses}

\paragraph{Controlled Comparison with Continuous Cropping.}
To disentangle the contribution of our training data from that of the tree-grounded action space, we train a continuous \texttt{crop\_video} agent (Table~\ref{tab:ablate_combined}) that is identical to \model~except for the action space. Specifically, this agent uses the same Qwen3-VL-8B backbone, training data, training budget, caption memory, and turn limit as \model. The agent is also trained on the same detour-and-recovery trajectories, where each recovery step is expressed as a crop back to the parent or sibling interval. Trained on our data, this agent achieves $15.5$ mIoU on CG-Bench, compared to $4.3$ mIoU of the published LongVT. On top of the identical data and captions, the tree-grounded action space further improves temporal grounding by $1.3$--$4.2$ points and QA accuracy by up to $9.8$ points. These results indicate that our training data contributes a large part of the improvement, while the tree-grounded action space brings consistent additional gains that do not come from caption quality.

\paragraph{Turn-Budget-Matched Comparison.}
To study whether the gains of \model~come from a longer search horizon, we construct LongVT-5, which calls \texttt{crop\_video} in the first four turns and answers in the fifth turn. As shown in Table~\ref{tab:turn_budget}, LongVT-5 improves over LongVT by only $0.2$ mIoU, while \model~outperforms LongVT-5 by $12.3$ mIoU with a similar number of turns. These results indicate that the gains of \model~mainly come from the tree-grounded action space rather than a longer search horizon.

\begin{table}[h]
\centering
\caption{\textbf{Turn-budget-matched comparison on CG-Bench.} Extending LongVT to five turns brings only marginal gains, while \model~outperforms it by a large margin with a similar number of turns.}
\label{tab:turn_budget}
\small
\begin{tabular}{lccc}
\toprule
Method & Turns & mIoU & Acc \\
\midrule
LongVT~\cite{yang2025longvt} & 2.0 & 4.3 & 17.4 \\
LongVT-5 & 5.0 & 4.5 & 18.3 \\
\rowcolor{gray!12}
\model~(Ours) & 4.8 & \textbf{16.8} & \textbf{36.4} \\
\bottomrule
\end{tabular}
\end{table}

\paragraph{Flat vs.\ Hierarchical Tree.}
In Table~\ref{tab:ablate_combined}, the flat tree achieves slightly higher T-F1 than the hierarchical tree on Haystack-Ego4D ($18.9$ vs.\ $18.4$). We note that the flat tree adaptively divides the video into short segments that are directly connected to the root. Since the clues in Haystack-Ego4D are uniformly short ($9.5$s on average), short segments are the right scale for almost every question, so direct access to these segments is well matched to this benchmark. In contrast, CG-Bench and Haystack-LVBench have more variable clue durations, and the hierarchical tree performs better on all their metrics.

\paragraph{Oracle Duration-Adaptive Tree.}
To study whether adapting the tree to the clue duration can bring further gains, we construct an oracle tree using the ground-truth clue duration, where nodes shorter than twice the clue duration are removed together with their descendants. As shown in Table~\ref{tab:oracle_tree}, the oracle tree achieves $16.2$ mIoU and $35.5$ accuracy, slightly lower than our default tree. We hypothesize that the agent already decides at each node whether to descend further or answer, so it can stop at the natural scale of the clue on its own.

\begin{table}[h]
\centering
\caption{\textbf{Comparison with an oracle duration-adaptive tree on CG-Bench.} Pruning the tree with the ground-truth clue duration does not improve performance.}
\label{tab:oracle_tree}
\small
\begin{tabular}{lcc}
\toprule
Tree structure & mIoU & Acc \\
\midrule
Oracle duration-adaptive tree & 16.2 & 35.5 \\
\rowcolor{gray!12}
Adaptive temporal segmentation (default) & \textbf{16.8} & \textbf{36.4} \\
\bottomrule
\end{tabular}
\end{table}

\paragraph{Incorporating Shot Boundary Detection.}
To study whether shot boundary detection can further improve adaptive temporal segmentation, we incorporate OmniShotCut~\cite{wang2026omnishotcut} into our tree construction. Specifically, we keep our hierarchical tree but snap each boundary to the nearest shot boundary detected by OmniShotCut. As shown in Table~\ref{tab:shot_boundary}, this variant slightly decreases mIoU ($16.3$ vs.\ $16.8$) with the same accuracy. These results indicate that aligning our boundaries with editing cuts provides no additional benefit.

\begin{table}[h]
\centering
\caption{\textbf{Incorporating shot boundary detection on CG-Bench.} Snapping our boundaries to the nearest OmniShotCut shot boundary does not improve performance.}
\label{tab:shot_boundary}
\small
\begin{tabular}{lcc}
\toprule
Tree construction & mIoU & Acc \\
\midrule
Snapped to OmniShotCut~\cite{wang2026omnishotcut} boundaries & 16.3 & \textbf{36.4} \\
\rowcolor{gray!12}
Adaptive temporal segmentation (default) & \textbf{16.8} & \textbf{36.4} \\
\bottomrule
\end{tabular}
\end{table}

\paragraph{Visual Encoder for Segmentation.}
To study which visual encoder is more suitable for adaptive temporal segmentation, we conduct an ablation study with two widely used visual encoders, CLIP~\cite{clip} and DINOv2~\cite{oquab2023dinov2}, while keeping all other components unchanged. The results are shown in Table~\ref{tab:visual_encoder}. From the table, we can see that CLIP outperforms DINOv2 by $0.6$ mIoU and $1.3$ accuracy, possibly because its vision-language pretraining helps capture semantic transitions in the video. Meanwhile, DINOv2 still achieves comparable results, showing that \model~is robust to the choice of visual encoder.

\begin{table}[h]
\centering
\caption{\textbf{Effect of the visual encoder for adaptive temporal segmentation on CG-Bench.} CLIP slightly outperforms DINOv2, and \model~is robust to the choice of visual encoder.}
\label{tab:visual_encoder}
\small
\begin{tabular}{lcc}
\toprule
Visual encoder & mIoU & Acc \\
\midrule
DINOv2~\cite{oquab2023dinov2} & 16.2 & 35.1 \\
\rowcolor{gray!12}
CLIP (default) & \textbf{16.8} & \textbf{36.4} \\
\bottomrule
\end{tabular}
\end{table}

\paragraph{Captioning Model.}
To study how the quality of node captions affects \model, we conduct an ablation study with three captioning models of different sizes: Qwen3-VL-2B, Qwen3-VL-4B, and \model-8B (default). For all variants, we keep the \model-8B action model and the search configuration fixed. The results are shown in Table~\ref{tab:captioning_model}. Stronger captioning models lead to better performance, improving mIoU by $1.2$ and accuracy by $3.9$ from 2B to 8B.

\begin{table}[h]
\centering
\caption{\textbf{Effect of the captioning model on CG-Bench.} The action model is fixed to \model-8B. Stronger captioning models lead to better performance.}
\label{tab:captioning_model}
\small
\begin{tabular}{lcc}
\toprule
Captioning model & mIoU & Acc \\
\midrule
Qwen3-VL-2B & 15.6 & 32.5 \\
Qwen3-VL-4B & 15.9 & 33.1 \\
\rowcolor{gray!12}
\model-8B (default) & \textbf{16.8} & \textbf{36.4} \\
\bottomrule
\end{tabular}
\end{table}

\paragraph{Runtime Analysis.}
Table~\ref{tab:runtime_breakdown} reports the runtime of each component of \model~on CG-Bench. Most of the runtime is spent on action generation and memory construction (mostly caption generation), while CLIP segmentation accounts for only $14\%$ of the total runtime. We further compare the peak GPU memory in Table~\ref{tab:peak_memory}. \model~uses $20.0$ GB of peak GPU memory, $34\%$ less than Qwen3-VL-8B with $384$ uniformly sampled frames ($30.5$ GB), since \model~processes at most $64$ frames at each turn. \model-fast has the same peak GPU memory as \model.

\begin{table}[h]
\centering
\caption{\textbf{Runtime breakdown of \model~on CG-Bench.} We report the average runtime per video on a single H100 GPU.}
\label{tab:runtime_breakdown}
\small
\begin{tabular}{lc}
\toprule
Component & Runtime (s) \\
\midrule
CLIP segmentation & 8.8 \\
Action generation & 24.7 \\
Memory construction & 28.1 \\
\midrule
Total & 61.6 \\
\bottomrule
\end{tabular}
\end{table}

\begin{table}[h]
\centering
\caption{\textbf{Peak GPU memory on CG-Bench.} We report the peak GPU memory per video on a single H100 GPU.}
\label{tab:peak_memory}
\small
\begin{tabular}{lc}
\toprule
Method & Peak GPU memory (GB) \\
\midrule
Qwen3-VL-8B ($384$ frames)~\cite{qwen3vl} & 30.5 \\
\model-fast (Ours) & \textbf{20.0} \\
\rowcolor{gray!12}
\model~(Ours) & \textbf{20.0} \\
\bottomrule
\end{tabular}
\end{table}

\section{More Qualitative Results}
\label{sec:visualization}

\begin{figure}[h]
  \centering
  \includegraphics[width=\linewidth]{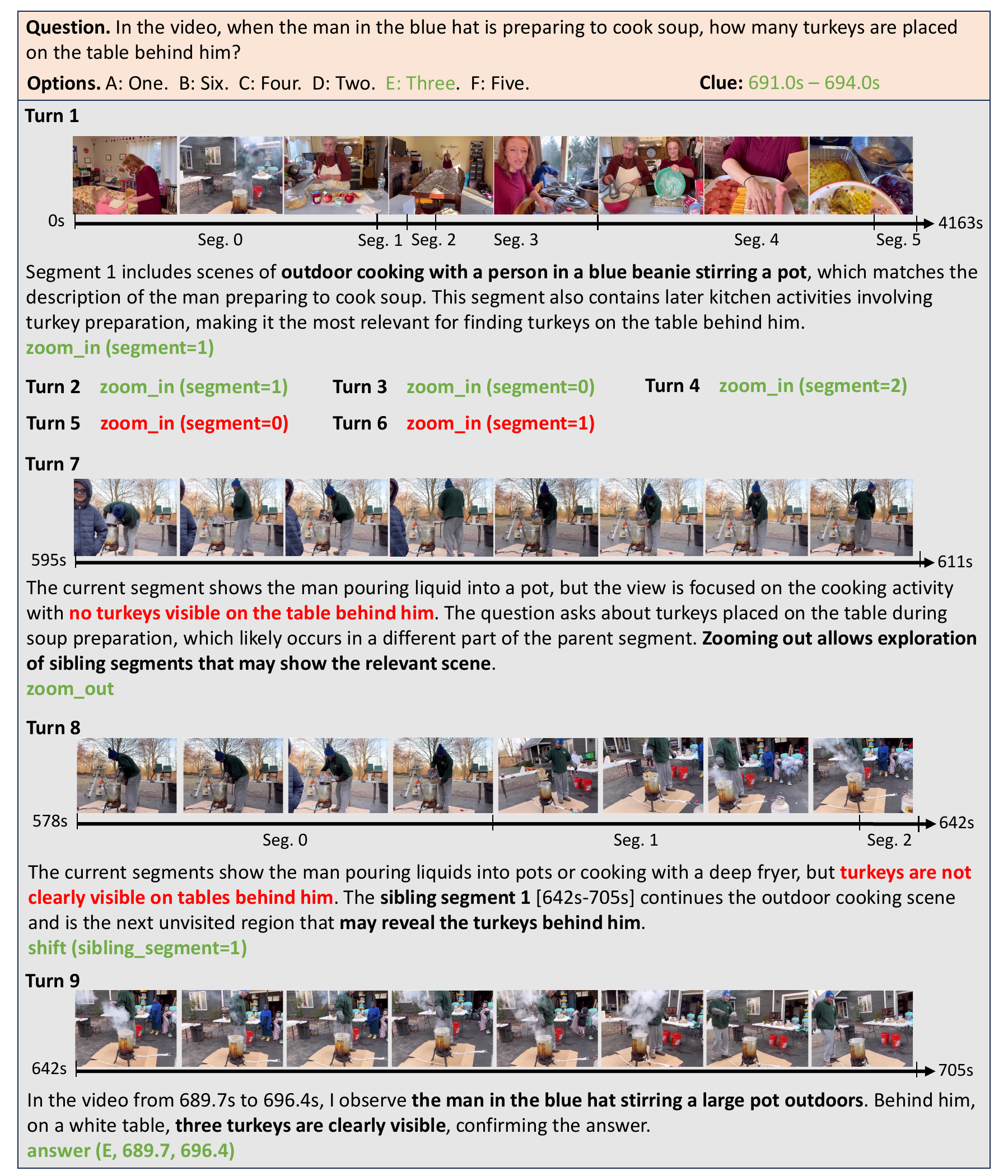}
  \caption{\textbf{Qualitative example: recovery via \texttt{zoom\_out} and \texttt{shift}.} The question asks for the number of turkeys on the table behind the man. The agent first follows several \texttt{zoom\_in} steps into an incorrect branch that shows the man but no turkeys (turns~1–6). It then recovers by using \texttt{zoom\_out} to ascend to the parent node (turn~7), followed by \texttt{shift} to move laterally to an unvisited sibling (turn~8). This sibling contains the relevant evidence—turkeys on the table behind the man—allowing the agent to answer correctly (turn~9). This trajectory illustrates how \texttt{zoom\_out} and \texttt{shift} work together to recover from incorrect exploration.}
  \label{fig:qual3}
\end{figure}

\begin{figure}[h]
  \centering
  \includegraphics[width=\linewidth]{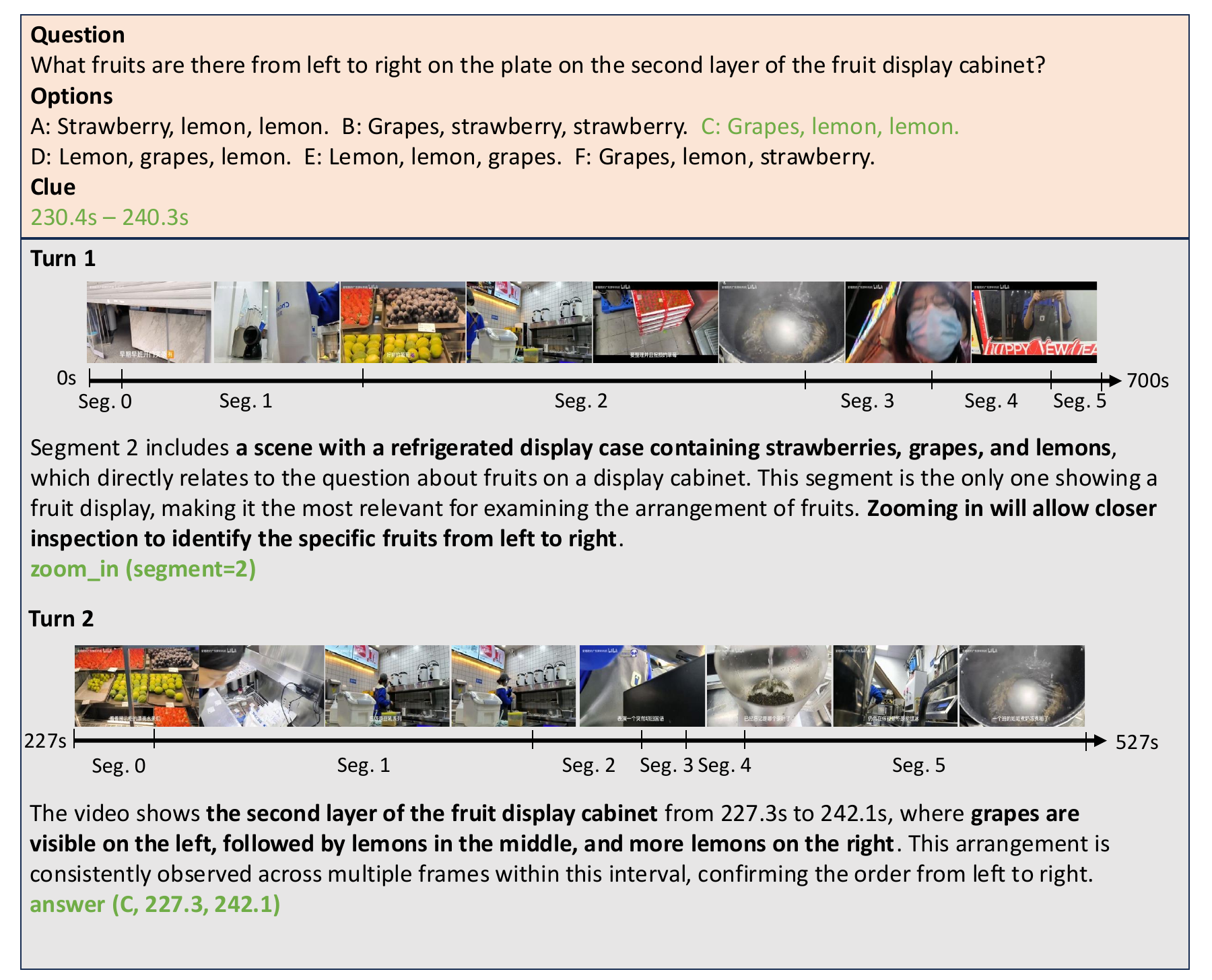}
  \caption{\textbf{Qualitative example: hierarchical \texttt{zoom\_in} without backtracking.} The question asks about the types of fruits placed on the cabinet. The agent directly reaches the relevant segment containing both the fruits and the cabinet after a single \texttt{zoom\_in} step (turn~1), and then terminates with \texttt{answer} in the next turn (turn~2).}
  \label{fig:qual1}
\end{figure}

\paragraph{Qualitative Examples of Tree Navigation.}
Figures~\ref{fig:qual3}–\ref{fig:qual1} present two additional qualitative results produced by VTS. Figure~\ref{fig:qual3} illustrates a deeper nine-turn trajectory that uses both backtracking primitives. The question asks how many turkeys are placed on the table behind the man. The agent first follows several \texttt{zoom\_in} actions into an incorrect branch, where the man is pouring liquid into a pot but no turkeys are visible on the table behind him. After recognizing this error, the agent issues \texttt{zoom\_out} to return to the parent node, followed by \texttt{shift} to move laterally to an unvisited sibling. This sibling contains both the turkeys on the table and the man, allowing the agent to terminate with \texttt{answer}.
Figure~\ref{fig:qual1} shows a short two-turn trajectory for a question asking about the types of fruits placed on the cabinet. In the first \texttt{zoom\_in} turn, the agent directly reaches the segment containing both the fruits and the cabinet, and then terminates with \texttt{answer} in the next turn. This represents the common case for questions whose evidence is visually salient at a coarse level, showing that the agent avoids unnecessary exploration when the initial descent is already correct.

\begin{figure}[h]
  \centering
  \includegraphics[width=\linewidth]{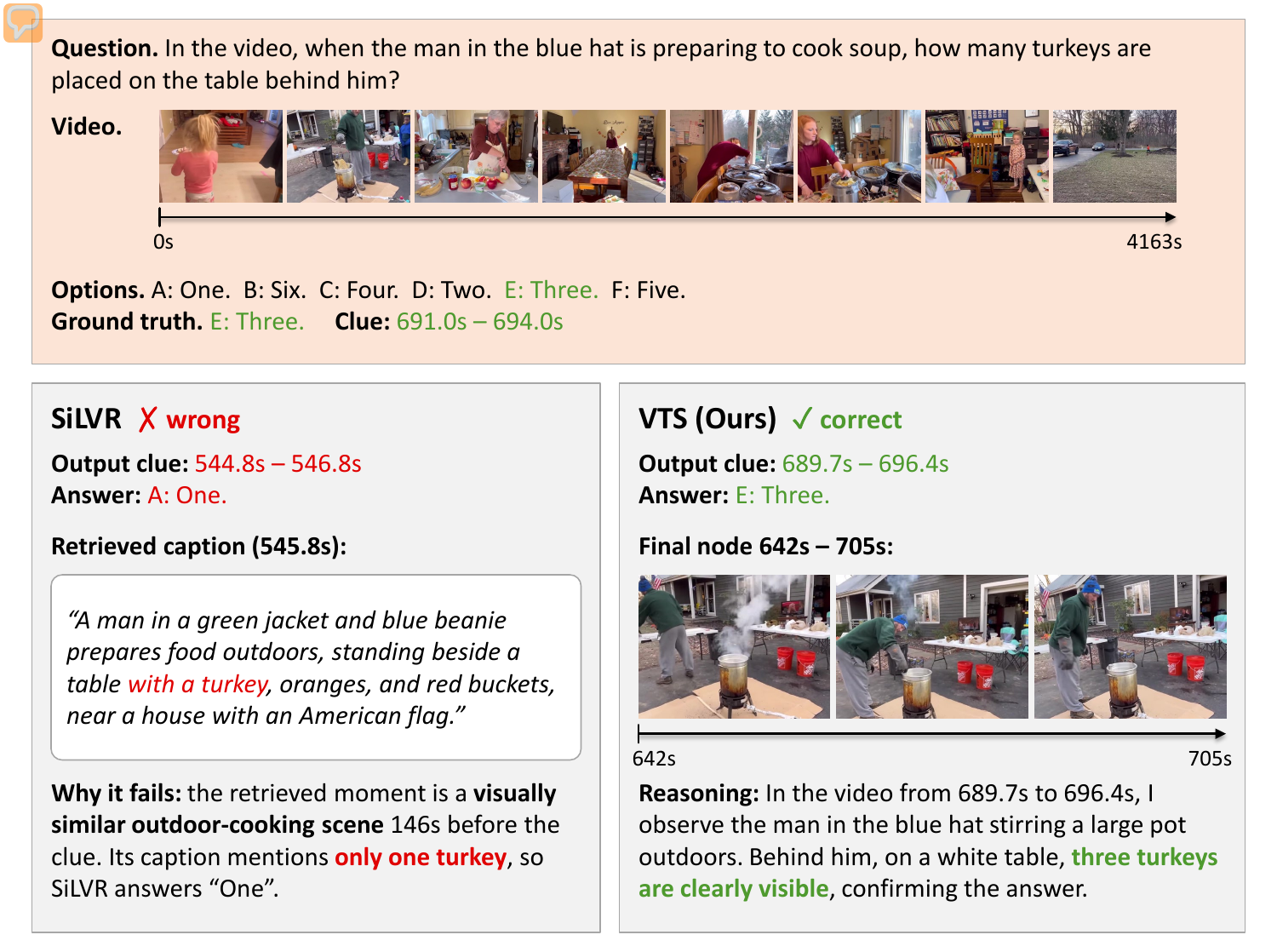}
  \caption{\textbf{Comparison between \model~and SiLVR.} The question asks how many turkeys are on the table behind the man in the blue hat; the ground-truth clue is 691.0s--694.0s in a 4163s video. SiLVR retrieves a visually similar outdoor-cooking moment 146s earlier (544.8s--546.8s) whose caption mentions a single turkey, and answers ``One''. \model~navigates to the node 642s--705s, observes three turkeys on the table behind the man, and answers ``Three'' with evidence 689.7s--696.4s.}
  \label{fig:qual_silvr}
\end{figure}

\paragraph{Comparison between \model~and SiLVR.}
Figure~\ref{fig:qual_silvr} compares \model~with SiLVR~\cite{zhang2025silvr} on a question that requires counting objects at a specific moment. SiLVR's caption-based reasoning returns a moment that matches the textual description of the question (a man in a blue beanie cooking outdoors beside a table with a turkey) but precedes the queried moment by $146$s. Because SiLVR reasons only over captions, it cannot verify the count and commits to ``One''. \model~instead reasons over the frames of the selected node and confirms the count visually before answering.

\begin{figure}[h]
  \centering
  \includegraphics[width=\linewidth]{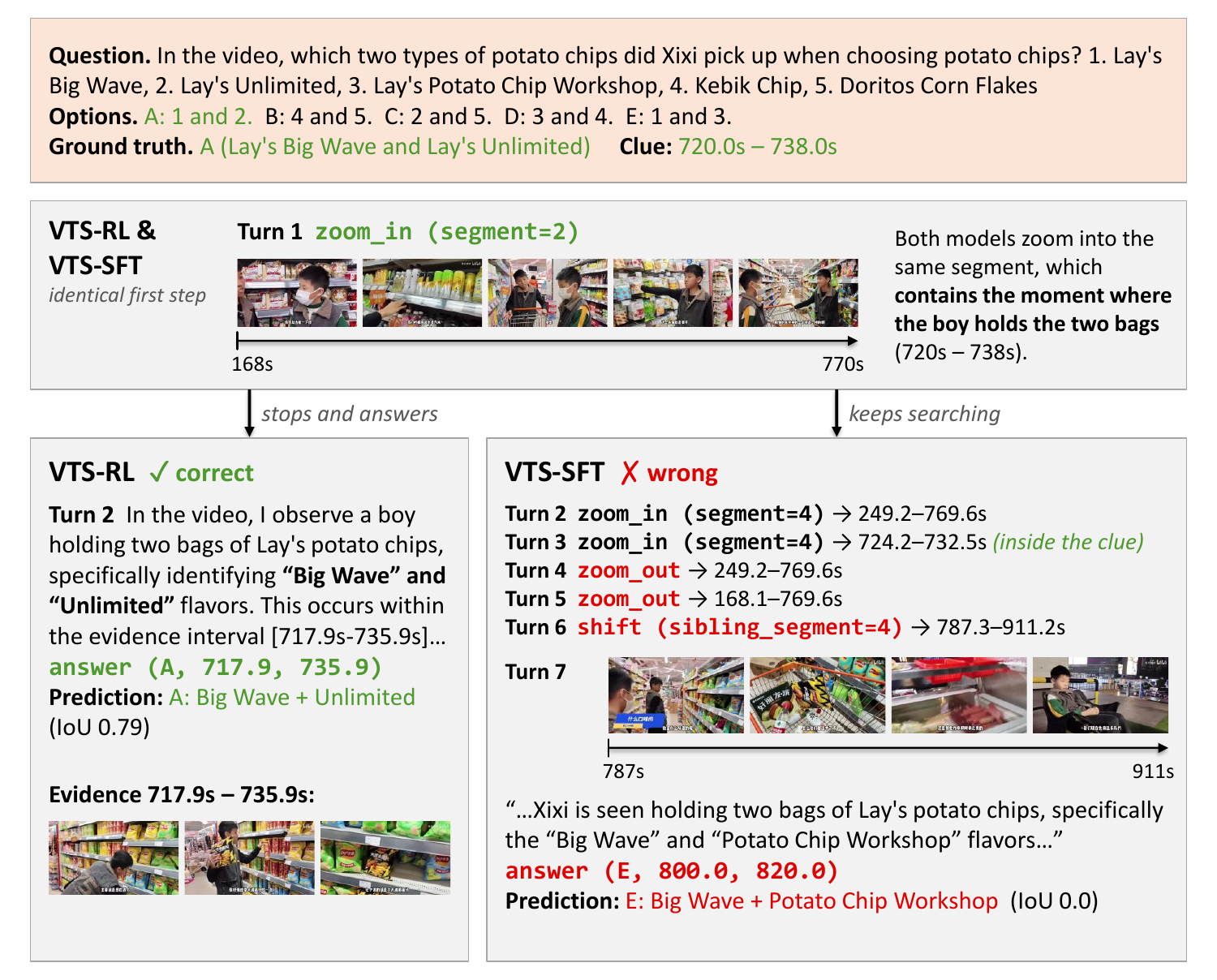}
  \caption{\textbf{Comparison between \model-RL and \model-SFT.} The question asks which two brands of potato chips the boy picks up; the ground-truth clue is 720.0s--738.0s. Both models first \texttt{zoom\_in} to the segment 168.1s--769.6s, which contains the evidence. \model-RL recognizes that the evidence is sufficient and answers correctly at turn~2 with evidence 717.9s--735.9s (IoU 0.79). \model-SFT keeps searching: it reaches a node inside the clue (724.2s--732.5s), but then issues \texttt{zoom\_out} twice, \texttt{shift}s to the sibling segment 787.3s--911.2s, and answers incorrectly from it (IoU 0.0).}
  \label{fig:qual_sft_rl}
\end{figure}

\paragraph{Comparison between \model-RL and \model-SFT.}
Figure~\ref{fig:qual_sft_rl} compares the trajectories of \model-RL and \model-SFT on the same question. Both models take the same first \texttt{zoom\_in}, which already reaches a segment containing the evidence. They differ in when they stop. \model-RL recognizes that the evidence is sufficient and answers correctly at turn~2. \model-SFT keeps searching. It zooms into a node that lies entirely inside the ground-truth clue ($724.2$s--$732.5$s), yet backtracks with two \texttt{zoom\_out} actions and \texttt{shift}s to a sibling segment that does not show the boy choosing the chips. There it misidentifies the second brand as ``Potato Chip Workshop'' and answers incorrectly. Both models use the same navigation actions, so the difference lies in deciding when the evidence is sufficient. SFT imitates demonstrated trajectories, whereas RL optimizes the whole trajectory for the answer and grounding outcome, which teaches the agent to terminate once it has found the evidence.

\begin{figure}[h]
  \centering
  \includegraphics[width=\linewidth]{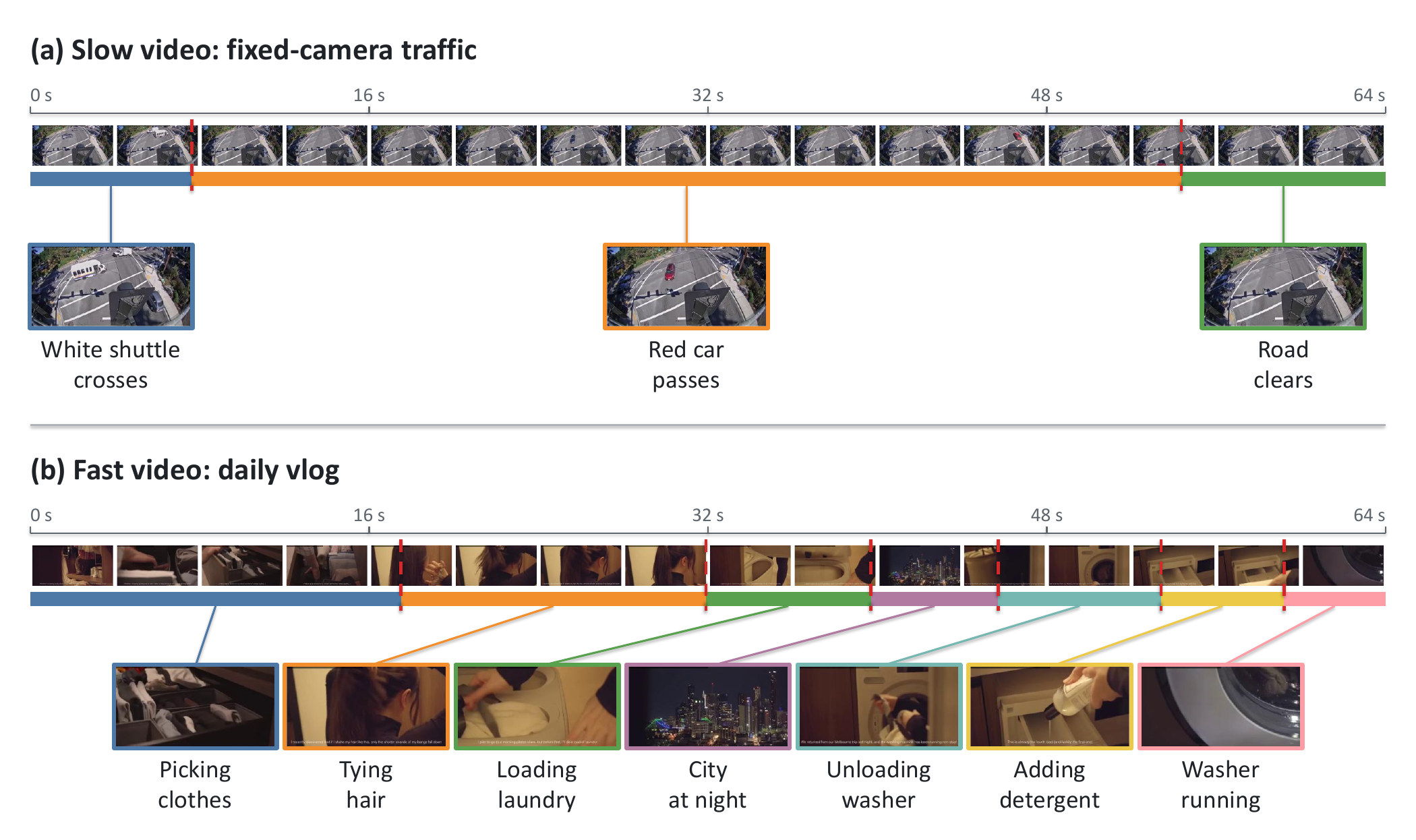}
  \caption{\textbf{Qualitative examples of adaptive temporal segmentation.} Each panel shows a 64-second clip with uniformly sampled frames (top), the resulting segments (colored bar) and boundaries (red dashed lines), and one representative frame per segment (bottom). (a) A fixed-camera traffic video with no editing cuts is split into three segments, the minimum allowed, with boundaries where a white shuttle finishes crossing the intersection and where a red car passes. (b) A daily-life vlog with frequent scene changes is split into seven segments, each corresponding to a distinct activity, such as picking clothes, tying hair, and loading laundry.}
  \label{fig:seg_examples}
\end{figure}

\paragraph{Qualitative Examples of Temporal Segmentation.}
Figure~\ref{fig:seg_examples} shows adaptive temporal segmentation on two $64$-second clips, one that changes slowly and one that changes rapidly. When a video changes gradually, few consecutive-frame distances exceed the adaptive threshold. In this case we add the next-largest distances until the node has at least $3$ children, so every node is split along its most salient changes. The fixed-camera traffic clip in Figure~\ref{fig:seg_examples}(a) has no editing cuts and is split into only three segments, with boundaries where a white shuttle finishes crossing the intersection and where a red car passes. When a video changes frequently, many distances exceed the threshold. We keep only the $7$ strongest boundaries, which caps each node at $8$ children and keeps the branching factor bounded. The vlog clip in Figure~\ref{fig:seg_examples}(b) spans the same $64$ seconds but is split into seven segments, each corresponding to a distinct activity, such as picking clothes, tying hair, or loading laundry. Moreover, visually subtle evidence does not need to produce a segment boundary, since the \texttt{answer} action predicts the evidence interval from the frames of the current segment.

\section{Limitations and Future Work}
\label{sec:limitations}
\model~has several limitations. The current training data provides a single evidence interval per question, so \model~is trained to output one interval. Extending \model~to questions whose evidence is distributed across multiple disjoint temporal regions would require multi-interval annotations and corresponding training trajectories. The adaptive temporal segmentation relies on CLIP embeddings to detect visual and semantic transitions. Boundary detectors designed for long-form video could yield finer trees and additional gains. Finally, trajectory synthesis depends on external vision-language and language models for captioning and reasoning, so the quality of the training signal is bounded by these models. We expect it to improve as stronger open-source models become available.